\documentclass[twoside,11pt]{article}

\usepackage{blindtext}

\usepackage[abbrvbib, preprint]{jmlr2e}
\usepackage{amsmath}
\usepackage{dsfont}
\usepackage{mathtools}
\usepackage{bm}
\usepackage{graphicx}
\usepackage{subcaption}
\usepackage{multirow}
\usepackage{url}
\usepackage[thinc]{esdiff}
\usepackage{comment}
\usepackage{xcolor}
\usepackage[createShortEnv]{proof-at-the-end}
\usepackage[ruled]{algorithm2e}
\usepackage{rotating}
\usepackage{pdflscape}

\usepackage{caption, threeparttable}

\newcommand{\indep}{\perp \!\!\! \perp}
\newcommand{\notindep}{\not\!\perp\!\!\!\perp}

\newcommand{\beginsupplement}{%
        \setcounter{table}{0}
        \renewcommand{\thetable}{S\arabic{table}}%
        \setcounter{figure}{0}
        \renewcommand{\thefigure}{S\arabic{figure}}%
     }

\usepackage{lastpage}
\jmlrheading{23}{2024}{1-\pageref{LastPage}}{1/21; Revised 5/22}{9/22}{21-0000}{Marco Simnacher, Xiangnan Xu, Hani Park, Christoph Lippert, Sonja Greven}

\ShortHeadings{Deep Nonparametric Conditional Independence Tests}{Simnacher, Xu, Park, Lippert and Greven}
\firstpageno{1}

\begin{document}

\title{Deep Nonparametric Conditional Independence Tests for Images}

\author{\name Marco Simnacher \email marco.simnacher@hu-berlin.de \\
\name Xiangnan Xu \email xiangnan.xu@hu-berlin.de \\
       \addr Chair of Statistics\\
       Humboldt-Universität zu Berlin\\
       Spandauer Str. 1, 10178 Berlin, Germany
       \AND
       \name Hani Park \email hani.park@hpi.de \\\name Christoph Lippert \email Christoph.Lippert@hpi.de \\
       \addr Chair Digital Health \& Machine Learning\\
       Hasso-Plattner-Institut for Digital Engineering\\
       Prof.-Dr.-Helmert-Straße 2-3, 14482 Potsdam, Germany
        \AND
       \name Sonja Greven \email sonja.greven@hu-berlin.de\\
       \addr Chair of Statistics\\
       Humboldt-Universität zu Berlin\\
       Spandauer Str. 1, 10178 Berlin, Germany}
       
\editor{My editor}

\maketitle

\begin{abstract}
Conditional independence tests (CITs) test for conditional dependence between random variables. As existing CITs are limited in their applicability to complex, high-dimensional variables such as images, we introduce deep nonparametric CITs (DNCITs). The DNCITs combine embedding maps, which extract feature representations of high-dimensional variables, with nonparametric CITs applicable to these feature representations. For the embedding maps, we derive general properties on their parameter estimators to obtain valid DNCITs and show that these properties include embedding maps learned through (conditional) unsupervised or transfer learning. For the nonparametric CITs, appropriate tests are selected and adapted to be applicable to  feature representations. Through simulations, we investigate the performance of the DNCITs for different embedding maps and nonparametric CITs under varying confounder dimensions and confounder relationships. We apply the DNCITs to brain MRI scans and behavioral traits, given confounders, of healthy individuals from the UK Biobank (UKB), confirming null results from a number of ambiguous personality neuroscience studies with a larger data set and with our more powerful tests. In addition, in a confounder control study, we apply the DNCITs to brain MRI scans and a confounder set to test for sufficient confounder control, leading to a potential reduction in the confounder dimension under improved confounder control compared to existing state-of-the-art confounder control studies for the UKB. Finally, we provide an R package implementing the DNCITs.
\end{abstract}


\section{Introduction}

A Conditional independence test (CIT) is a statistical test for conditional dependence between two random variables, $X$ and $Y$, given a third variable, $Z$. That is, a CIT is designed to test the null hypothesis of conditional independence, $H_0:X\indep Y|Z$, against the alternative hypothesis of conditional dependence, $H_1:X\notindep Y|Z$. $X$ and $Y$ are conditionally dependent, given $Z$, if $X$ (or $Y$) provides additional information about $Y$ (or $X$), given $Z$. To test the hypothesis, a CIT typically consists of a test statistic that measures the conditional dependence between $X$ and $Y$, given $Z$, and a comparison of the observed test statistic with its distribution under the null hypothesis of conditional independence (CI). 
A CIT should be valid and control the type 1 error (T1E), i.e. the probability of falsely rejecting the null hypothesis, while simultaneously achieving high power, i.e. a high probability of correctly rejecting the null hypothesis of CI. 

CITs are used for causal discovery, graphical model learning and feature selection because of the connection between CI and causal inference, prediction sufficiency, and parameter identification \citep{dawid1979conditional}. They are currently used in a variety of fields, including biomedical data \citep{bellot2019conditional, li2022searching, katsevich2022power}, environmental data \citep{runge2018conditional, runge2023causal}, neuroimaging data \citep{grosse2016identification}, and economic data \citep{huang2016flexible}, where datasets are often multimodal and consist of complex objects such as images. 

A limitation of existing CITs, which we discuss throughout the paper, is their focus on low-dimensional, often univariate $X$ and $Y$. Such CITs cannot easily be applied to complex, high-dimensional objects because they are theoretically or computationally inapplicable, do not hold the T1E, or lack power against common alternatives. In particular, their test statistics or null hypothesis comparisons are often inapplicable because they are limited to low-dimensional $X$ and $Y$, either theoretically or computationally. When the CITs are applicable, their T1Es are often inflated by uncontrolled confounding because their comparison to the null hypothesis is typically inappropriate for complex, high-dimensional $X$ or $Y$ with nonlinear relationships with $Z$. Finally, even if the CITs control the T1E at the specified significance level, they lack power against reasonable alternatives, because their test statistics do not account for the nonlinear relationship between $X$ and $Y$ given $Z$, although this is common for complex, high-dimensional $X$ or $Y$.

Based on these observations, we see three requirements for the proposed deep nonparametric conditional independence tests (DNCITs): (i) Tests must be applicable to complex, high-dimensional random variables, given potential vector-valued confounders. (ii) Tests must control the T1E, even in settings with strong and nonlinear confounding. (iii) Tests should have high power against large sets of alternatives, especially for nonlinear relationships among all random variables combined with small effect sizes. The first requirement is unique to our context, while requirements two and three are standard for (nonparametric) statistical testing, especially nonparametric CI testing \citep{lehmann1986testing, shah2020hardness}. In addition to these requirements, we focus on images $X$ and scalars $Y$ throughout this paper due to their wide availability, although the DNCITs can be applied to general complex, high-dimensional $X$ and $Y$ as shown in Appendix \ref{app:CIT_after_transformations}.

\begin{figure}
    \centering
    \includegraphics[width=0.5\textwidth]{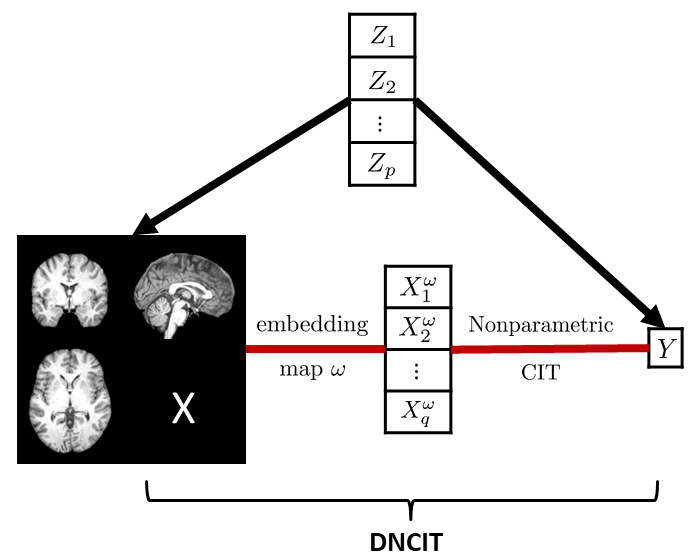}
    \caption{The DNCIT for an image $X$, a scalar $Y$ and a vector-valued confounder $Z=(Z_1,\hdots,Z_p)$. The black arrows indicate causal effects from $Z$ to $X$ and $Y$, the red lines represent the two steps of the DNCITs. DNCITs test for conditional dependence between $X$ and $Y$ by mapping $X$ through an embedding map $\omega$ to a vector-valued feature representation $X^\omega=(X^\omega_1,\hdots,X^\omega_q)$ in step one and applying a nonparametric CIT which tests for conditional dependence between $X^\omega$ and $Y$ given $Z$ in step two.}
    \label{fig:cit_setting}
\end{figure}

To meet these requirements, the proposed DNCITs consist of two steps (Figure \ref{fig:cit_setting}): First, they map the images to vector-valued feature representations $X^\omega$ using embedding maps $\omega$. Second, a suitable nonparametric CIT tests for conditional associations between $X^\omega$ and $Y$. The dimension reduction of the embedding maps makes the DNCITs applicable to an image and a scalar. T1E control is ensured through conditionally independently learned embedding maps and the nonlinear confounder control of the nonparametric CITs. Finally, the resulting DNCITs have high power to detect nonlinear relationships between $X$ and $Y$ through the nonlinearities encoded in the embedding maps together with the test statistics of the nonparametric CITs. Note that our approach is a framework not tied to any particular choice of embedding map and CIT, and could thus also profit from further advances in both areas, while we discuss suitable choices for both both theoretically as well as based on simulations in this paper.

In summary, our approach addresses the limitations of existing CITs and provides a robust framework for CI testing for complex, high-dimensional data such as images. Our contributions include the following:
\begin{enumerate}
\item Introduction of DNCITs based on kernels, conditional permutation, conditional mutual information, and prediction models, applicable to an image and a scalar, potentially given a vector-valued confounder.
\item Derivation of theoretical results on the validity of the DNCITs and the T1E excess of one specific DNCIT.
\item Introduction of a simulation design with real-world confounding effects on the images and controllable relationship between $X,Z$ and $Y$.
\item Evaluation of the T1E and power of DNCITs for different embedding maps and nonparametric CITs, for different confounder dimension and confounder relationship, in a simulation study for UKB brain MRI scans.
\item Application of DNCITs in two real-world settings: examining the dependence between brain MRI scans and behavioral traits while accounting for confounders, and testing for sufficient confounder control for brain MRI scans in the UKB, extending recent findings in the literature.
\item Implementation of all DNCITs in the publicly available R-package DNCIT.
\end{enumerate}

The paper is organized as follows: in section \ref{sec:litRev}, we review the relevant literature on nonparametric CITs and statistical tests with DL embedding maps. The general framework of DNCITs is introduced in section \ref{sec:DNCIT_general}. We first present theoretical results on the validity of the DNCITs, then derive theoretical conditions for embedding maps that lead to valid DNCITs and study the effect of these embedding maps on the power of the tests. Finally, we discuss  nonparametric CITs for application to the feature representations. Simulation results and two applications are presented in the sections \ref{sec:sim} and \ref{sec:applications}, respectively. Finally, section \ref{sec:conclusion} concludes.

\section{Review of Relevant Literature}\label{sec:litRev}

We first discuss current work on nonparametric CITs with respect to their applicability to vector-scalar-valued data, which we build on for the second step of the proposed DNCITs. We then review the literature on combining statistical tests with DL embedding maps, related to the first step of our DNCITs.

\subsection{Nonparametric Conditional Independence Tests}\label{subsec:nonparam_CIT_lit_review}

The often complex, nonlinear relations between the image $X$, $Y$, and $Z$ can result in complex, nonlinear relations between the feature representation $X^\omega$, $Y$, and $Z$ for general embedding maps in the DNCITs. 
For such complex relations between $X^\omega$, $Y$, and $Z$, nonparametric CITs allow DNCITs to have power while controlling for T1E. 
However, nonparametric CI testing has been shown to be a hard problem for continuous confounders in the sense that they only have power against certain subsets of alternatives and can only control the T1E for certain subsets of the null hypothesis, without making additional assumptions \citep{shah2020hardness, neykov2021minimax}.
That said, many nonparametric CITs with different assumptions have been introduced with a review given in \citet{li2020nonparametric}, and we here discuss and select suitable nonparametric CITs for the DNCIT framework.
We categorize nonparametric CITs according to their test statistic and comparison with the null hypothesis. 
To compare the performance of DNCITs using different CITs, we select CITs from different categories that allow transferability to our setting, i.e., that are applicable to a vector-valued $X^\omega$ and a scalar $Y$, can be computed for potentially continuous vector-valued $Z$, and are implemented. 
The included CITs are further discussed in subsection \ref{subsec:nonparam_CITs} and Appendix \ref{app:online}.

\subsubsection{Test Statistics}

The test statistic is typically a measure of the conditional dependence or deviation from CI, thereby affecting the power and T1E of the DNCITs through the kind of nonlinear relationships between $X^\omega$, $Y$, and $Z$ detectable in the measure. There are nonparametric CITs with test statistics based on kernels, regression residuals, predictions, and metrics between distributions.

\textbf{Kernel-based CITs} often simplify the problem of CI testing by restricting the function spaces considered within the test statistic to reproducing kernel Hilbert spaces (RKHSs). 
The test statistics of the Kernel CIT \citep[KCIT;][]{zhang2012kernel} and the Randomized conditional Correlation Test \citep[RCoT;][]{strobl2019approximate} measure the correlation of the residual functions after mapping $X^\omega, Y$ and $Z$ into RKHSs and regressing the respective mappings on mappings of $Z$. The test statistic of
\citet{doran2014permutation} uses the maximum mean discrepancy (MMD) between the observed and conditionally permuted mappings of $(X^\omega, Y, Z)$, while
\citet{zhang2023conditional}'s test statistic is based on a MMD between the joint and the product of the marginal distributions of $X^\omega$ and $Y$.  
In addition, \citet{huang2022kernel} propose the kernel partial correlation (KPC) estimated using kernels and geometric graph functionals as an extension of the conditional dependence measure introduced in \citet{azadkia2021simple}, which has been studied regarding its power in \cite{shi2021azadkia} and \citet{lin2023boosting}. 
Often, these kernel-based test statistics rely on rather weak assumptions regarding their equivalence to CI in their null hypothesis compared to other CITs, resulting in a potentially better T1E control. Additionally, they are often flexible in the dimensions of $X^\omega$, $Y$, and $Z$. However, their computation can be slow \citep[sec.~3.1.4]{li2020nonparametric}.
Thus, we select the RCoT due to its fast and stable implementation. Furthermore, the KPC is selected because of its stable implementation and the possibility of rather general domains for $X^\omega$, $Y$ and $Z$ . 

{\bf Residual-based CITs} test for the dependence of the residuals of the regression functions of $Y$ on $Z$ and $X^\omega$ on $Z$. 
\citet{shah2020hardness} propose the Generalized Covariance Measure (GCM), extended in \citet{scheidegger2022weighted, lundborg2022projected} and \citet{lundborg2022conditional}, which uses a normalized sum of the product of these residuals as the test statistic. 
Concurrently with the kernel-based RCoT, which also belongs to this category, \citet{zhang2017causal} apply also kernel-based transformations to $X^\omega$ and $Y$ followed by a kernel ridge regression  on $Z$, but using a permutation-based approach for the null comparison.
In \citet{duong2022conditional}, the residuals are obtained using conditional normalizing flows for one-dimensional $X^\omega$ and $Y$. 
These residual-based test statistics mostly rely on the fit of a regression model, which is often a fairly feasible task. However, they often assume low-dimensional $X^\omega$ and $Y$, which can make them computationally or even theoretically inapplicable to our setting of high-dimensional $X^\omega$. 
Since the RCoT makes the problem feasible via reducing the dimension of $X^\omega$ by mapping it to a lower dimensional feature space and then applying fast linear ridge regression to obtain the residuals, we select it for this category of CITs as well.

{\bf Prediction-based CITs} test for a significant increase in the accuracy 
of a prediction of $Y$ using both $X^\omega$ and $Z$ compared to using only $Z$ without the information in $X^\omega$. 
To remove the information in $X^\omega$, the Fast Conditional Independence Test \citep[FCIT;][]{chalupka2018fast} uses decision trees and the Predictive Conditional Independence Test \citep[PCIT;][]{burkart2017predictive} uses an aggregation over multiple prediction models without $X^\omega$, while the CIT based on the Conditional Predictive Impact \citep[CPI;][]{watson2021testing} computes the risk over arbitrary losses and \citet{spisak2022statistical} $R^2$ with conditionally independently generated samples $X^{\omega,(m)}$.
Similar, model agnostic prediction-based approaches can also be found in the conditional mean dependence literature \citep{fisher2019all, covert2021explaining, williamson2021nonparametric, cai2022test, dai2022significance, lundborg2022projected, williamson2023general}.
Unlike residual-based approaches, where $X^\omega$ appears as output, prediction-based approaches have the advantage of considering the high-dimensional $X^\omega$ as predictors in regression models. In addition, they do not require the same dimension for $X^\omega$ and $Y$ as many other CITs do. 
However, the tests often consider only certain parts of the conditional distribution of $Y$ given $X^\omega$ and $Z$ such as the mean, instead of the entire distribution, resulting in the weaker conditional mean dependence tests and a potential loss of power. 
We consider the FCIT as representative of prediction-based CITs, since it is explicitly designed and implemented as CIT.  

In contrast, {\bf Metric-based approaches}, which aim to characterize the CI by a metric between the joint and the product of the corresponding marginal distributions of $Y$ and $X^\omega$ given $Z$, consider the entire conditional distribution.
The Conditional Distance Independence Test \citep{wang2015conditional} achieves this by relying on conditional characteristic functions. 
Similar metric-based approaches have been proposed in \citet{su2007consistent, su2014testing, huang2016flexible} and \citet{runge2018conditional}. Since they consider the entire distribution of $X^\omega$,$Y$ and $Z$, a major challenge of these test statistics is the need to estimate some variant of the conditional densities, which suffers from the curse of dimensionality \citep{li2020nonparametric}. As a representative of metric-based CITs we select the Conditional Mutual Information (CMI) k-nearest-neighbor (knn) (CMIknn) CIT of \citet{runge2018conditional} due to its flexible, stable implementation, which aims to characterize the CI using the conditional mutual information as a test statistic.

\subsubsection{Comparison to the Null Hypothesis}

A CIT's T1E control relies on comparing the observed test statistic to its distribution under the null hypothesis.  A DNCIT inherits the T1E control from the T1E control of its nonparametric CIT for appropriately chosen embedding maps. There are null comparisons based on (asymptotic) distributions of the test statistic, local permutation, or conditional randomization.

For some test statistics an {\bf (asymptotic) distribution} under the null hypothesis can be derived. The null hypothesis can then be rejected if the test statistic exceeds the corresponding quantile of the distribution. Among others, the CITs of \citet{zhang2012kernel, strobl2019approximate, shah2020hardness} and \citet{scetbon2022asymptotic} are based on asymptotic distributions of the corresponding test statistics. This is often preferable in terms of the computational cost of the test, but may suffer from inflated T1Es for small sample sizes if the distribution can only be derived asymptotically. In this category fall the selected RCoT and FCIT.

{\bf Local permutation approaches} alternatively focus on locally permuting either $X^\omega$ or $Y$ or both. \citet{neykov2021minimax} and \citet{kim2022local} extend local permutation from discrete to continuous confounders by binning $Z$ and permuting $Y$ within these bins. Alternatively, the Classifier CIT \citep[CCIT;][]{sen2017model} permutes the values of $Y$ for the 1-nearest neighbor in $Z$, approximately resulting in samples from the null hypothesis. 
Similarly, the CMIknn \citep{runge2018conditional} randomly permutes $Y$ with one of its knns in $Z$.
Although these approaches are able to obtain samples which are approximately CI through simple knn and binning models, they can be computationally expensive and suffer from the curse of dimensionality in the confounder.
Due to its computational advantages and potential to optimize the power and T1E control depending on the number of knn, the CMIknn is selected for this category.
 
The {\bf conditional randomization test} \citep[CRT;][]{barber2015controlling, candes2018panning} requires knowledge on or an approximation of the marginal conditional distributions $\mathbb{P}^{X^{\omega}|Z}$ or $\mathbb{P}^{Y|Z}$ to compare the observed test statistic with  test statistics at generated samples from the marginal conditional distributions. 
\citet{katsevich2022power} and \citet{wang2022high} analyse the power of CRT-based statistical tests theoretically.  
\citet{shi2021double} and \citet{bellot2019conditional} propose CRT-based CITs approximating the marginal conditional distributions by Generative Adversarial Networks (GANs). 
Furthermore, the conditional permutation test \citep[CPT;][]{berrett2020conditional}, extended in \citet{spisak2022statistical}, was inspired by the CRT. It also requires additional knowledge about $\mathbb{P}^{X^{\omega}|Z}$ or $\mathbb{P}^{Y|Z}$, but uses a permutation of the observed sample. CRT and CPT provide T1E control, flexibility in the choice of the test statistic and thus, potentially large power for well chosen test statistics.
Due to its superior performance compared to the CRT with respect to the excess T1E \citep{berrett2020conditional}, the CPT together with the KPC as test statistic is selected. 

\subsection{Statistical Tests with Embedding Maps}

Statistical tests using latent representations typically involve projecting complex, high-dimensional objects onto lower-dimensional feature representations, followed by applying significance tests to these projections. Existing research has primarily focused on either two-sample testing and representation learning or real-world applications of CITs using specific embedding maps.

Specifically, \citet{kirchler2020two} propose two two-sample tests based on the Maximum Mean Discrepancy (MMD) between the feature representations of two high-dimensional objects. Similarly, \citet{liu2020learning} enhance the power of a two-sample test by optimizing kernels that generate the feature representations, maximizing test power relative to the learned representations. This approach builds on \citet{sutherland2021generative}, which optimizes the power of MMD-based two-sample tests over the choice of kernel.

\citet{duong2022conditional} construct a CIT testing for dependence between the latent representations of conditional normalizing flows for two one-dimensional real-valued variables, given a d-dimensional real-valued variable. 
\citet{pogodin2022efficient} derive a conditionally independent feature representation of potentially high-dimensional, complex objects and remark on a potential future extension of this framework to statistical testing. 

More directly related to our work, \citet{kirchler2022transfergwas} develop a CIT for genome-wide association studies (GWAS), projecting images onto a few leading principal components of  feature representations learned using sample splitting or transfer learning, and using these representations in parametric linear genetic association tests. \citet{rakowski2024transfergwas} applied this test to UKB data in a large-scale GWAS. Likewise, \citet{kook2024algorithm} applied a conditional mean dependence test based on the projected covariance measure of \cite{lundborg2022projected} on three multimodal datasets, including one with pre-trained embeddings of chest x-rays. 

While the listed theoretical works combine embedding maps with statistical tests, they either do not consider CITs or do not consider high-dimensional, complex objects \citep{duong2022conditional}. Although the applied studies above demonstrate the utility of CITs for image-scalar-valued data, we in contrast propose a general and modular nonparametric CIT framework, provide requirements for embedding maps to obtain theoretically valid CITs,
as well as conduct a comprehensive simulation study exploring the effects of test statistics and embeddings used on the T1E rate and power.

\section{Deep Nonparametric Conditional Independence Tests}\label{sec:DNCIT_general}

Let $(\mathcal{X}, \mathcal{F}_\mathcal{X}), (\mathcal{Y}, \mathcal{F}_\mathcal{Y}), (\mathcal{Z}, \mathcal{F}_\mathcal{Z})$ be measurable spaces and $X,Y,Z$ be $(\mathcal{X}, \mathcal{F}_\mathcal{X})$-, $(\mathcal{Y}, \mathcal{F}_\mathcal{Y})$-, $(\mathcal{Z}, \mathcal{F}_\mathcal{Z})$-valued random variables with joint distribution $\mathbb{P}^{X,Y,Z}$. Moreover, let $\hat{\beta}$ be a $(\mathcal{B},\mathcal{F}_{\mathcal{B}})$-valued random variable representing parameters of  $\omega:\mathcal{X}\times \mathcal{B}\to \mathfrak{X},(X,\hat{\beta})\mapsto \omega(X,\hat{\beta})$, a $(\mathfrak{X},\mathcal{F}_\mathfrak{X})$-valued random variable defined through a measurable function $\omega$ of $(X,\hat{\beta})$, called embedding map. We denote the random representation of the random variable $X$ obtained through the embedding map by $X^\omega=\omega(X,\hat{\beta})$. 
The focus of this paper is on $\mathcal{X}$ as a high-dimensional space of images, $\mathcal{Y}\subseteq\mathbb{R}$ for a univariate continuous variable, $\mathcal{Z}\subseteq\mathbb{R}^p$ for a multivariate confounder (or $p=0$ for (unconditional) independence tests), and $\mathfrak{X}\subseteq \mathbb{R}^q$ as the space of feature representations. Moreover, several nonparametric CITs are also directly applicable to vector-valued $Y$, allowing for the corresponding DNCITs to be applied to vector-valued $Y$ as well. For extensions of the framework to complex, high-dimensional $Y$ compare Appendix \ref{app:CIT_after_transformations}.

The idea of DNCITs is to translate the problem of testing
\begin{align}\label{test_hypotheses_image}
    H_0:X \indep Y|Z\quad vs.\quad H_1:X\notindep Y|Z
\end{align}
to testing
\begin{align}\label{test_hypotheses_embedding}
    H_0^\omega:X^\omega \indep Y|Z\quad vs.\quad H_1^\omega:X^\omega\notindep Y|Z
\end{align}
with an existing nonparametric CIT. Therefore, we assume that $(X_i, Y_i, Z_i), i=1,\hdots,n$ are $n$ independently and identically distributed (i.i.d.) copies of $(X,Y,Z)$. Additionally, we define $X^{\omega}_i=\omega(X_i,\hat{\beta})$, and write $X^n=(X_1,\hdots,X_n),(X,Y)^n=(X_i,Y_i)_{i=1}^n$, and analogously for all other combinations of $X_i^\omega,X_i,Y_i,Z_i, i=1,\hdots,n$.  
For the samples with images and their feature representations we write $S=(X, Y, Z)^n$ and $S^\omega=(X^\omega, Y,Z)^n$, respectively. 
Furthermore, let the underlying probability spaces for the samples be $(\mathcal{S}, \mathcal{F}_\mathcal{S}, \mathbb{P}^S)$ and analogously $(\mathcal{S}^\omega, \mathcal{F}_\mathcal{S}, \mathbb{P}^{S^\omega})$.\footnote{More precisely, let the underlying probability spaces be $(\mathcal{S}, \mathcal{F}_\mathcal{S}, \mathbb{P}^S)=(\prod_{i=1}^n(\mathcal{X}_i\times \mathcal{Y}_i\times \mathcal{Z}_i), \bigotimes_{i=1}^n(\mathcal{F}_{\mathcal{X}_i}\otimes\mathcal{F}_{\mathcal{Y}_i}\otimes\mathcal{F}_{\mathcal{Z}_i}), \bigotimes_{i=1}^n\mathbb{P}^{(X_i,Y_i,Z_i)})=((\mathcal{X}\times\mathcal{Y}\times\mathcal{Z})^n, \mathcal{F}_{\mathcal{X}\times\mathcal{Y}\times \mathcal{Z}}^{\otimes n}, \mathbb{P}^{(X,Y,Z)^n})$ and analogously $(\mathcal{S}^\omega, \mathcal{F}_\mathcal{S}, \mathbb{P}^{S^\omega})=((\mathfrak{X}\times\mathcal{Y}\times\mathcal{Z})^n, \mathcal{F}_{\mathfrak{X}\times\mathcal{Y}\times \mathcal{Z}}^{\otimes n}, \mathbb{P}^{(X^\omega,Y,Z)^n})$.} 
Then, we denote the collection of null and alternative distributions of \eqref{test_hypotheses_image} by $\mathcal{P}_0=\{\mathbb{P}^{S}|
X\indep Y|Z\}$ and $\mathcal{P}_1=\{\mathbb{P}^{S}|X\notindep Y|Z\}$, respectively, and correspondingly of \eqref{test_hypotheses_embedding} by $\mathcal{P}_0^{\omega}=\{\mathbb{P}^{S^\omega}|X^\omega\indep Y|Z\}$ and $\mathcal{P}_1^{\omega}=\{\mathbb{P}^{S^\omega}|X^\omega\notindep Y|Z\}$, respectively.

Then, we propose the DNCIT as measurable function
\begin{align}\label{equ:DNCIT}
    \varphi_{n, \omega, \theta}: \mathcal{S}\to \{0, 1\}, S\mapsto\varphi_{n,\theta}^\omega\circ \Bar{\omega}(S)
\end{align}
where $\omega$ is the embedding map depending implicitly on an estimator $\hat{\beta}$ of parameters $\beta$, $\Bar{\omega}$ takes the whole sample $S$ as input and maps it onto $S^\omega$ through the embedding map, the measurable function $\varphi_{n,\theta}^\omega:  \mathcal{S}^\omega  \to \{0,1\},S^\omega\mapsto \varphi_{n, \theta}^\omega(S^\omega)$
is a nonparametric CIT with parameters $\theta$ testing \eqref{test_hypotheses_embedding}, and $\varphi_{n,\omega, \theta}(S)=0$, respectively 1, define decisions for $H_0$ and $H_1$, respectively, in \eqref{test_hypotheses_image}. In the following, we omit the dependence on $n$ and write $\varphi_{\omega, \theta}$ for the DNCIT and $\varphi_{\theta}^\omega$ for the corresponding nonparametric CIT. The test statistic for a DNCIT is defined as
\begin{align}\label{equ:test_stat_DNCIT}
    T_{\omega,\theta}: \mathcal{S}\to \mathbb{R},S\mapsto T_{\omega,\theta}(S)=T_\theta^\omega\circ\Bar{\omega}(S)
\end{align}
where $T_\theta^\omega: \mathcal{S}^\omega\to \mathbb{R}, S^\omega\mapsto T_\theta^\omega(S^\omega)$
is the test statistic of the nonparametric CIT $\varphi_{\theta}^\omega$.

There are two components that can be chosen in DNCITs: 1.) the embedding map $\omega$ of images onto feature representations depending on an estimator $\hat{\beta}$ of parameters $\beta$, 2.) the nonparametric CIT $\varphi_{\theta}^\omega$ testing \eqref{test_hypotheses_embedding} and consisting of its test statistic $T_\theta^\omega$, and the comparison to the null distribution of the test statistic $T_\theta^\omega$. 
To ensure the resulting test is valid, in subsection \ref{subsec:theory_DNCITs} we first present theoretical results on the relationship between the null hypotheses in \eqref{test_hypotheses_image} and \eqref{test_hypotheses_embedding}, as well as on the validity of a DNCIT given certain assumptions on the corresponding embedding map and nonparametric CIT. Subsection \ref{subsec:feature_representations} translates these theoretical results to the validity of DNCITs using embedding maps obtained from different learning procedures (assuming valid nonparametric CITs), 
and discusses the effect of these embedding maps on the DNCITs' T1E and power. Finally, potentially suitable nonparametric CITs for these embedding maps are studied in subsection \ref{subsec:nonparam_CITs} and Appendix \ref{app:online}. 

\subsection{Validity of Deep Nonparametric Conditional Independence Tests}\label{subsec:theory_DNCITs}

In the following, we show that correctness of the null hypothesis for image and scalar in \eqref{test_hypotheses_image} implies the corresponding for feature representation and scalar in \eqref{test_hypotheses_embedding}, under an assumption on the embedding map which ensures that it does not introduce conditional dependence. A version of this theorem for non-scalar $Y$ and their corresponding embedding maps can be found in Appendix \ref{app:CIT_after_transformations}.

\begin{theoremE}[Relation of null hypotheses $H_0$ and $H_0^\omega$][end, text link={All proofs are provided in Appendix \ref{app:proofs}.}]\label{thm:null_hypotheses} 
Let $\hat{\beta}$ be a $(\mathcal{B},\mathcal{F}_\mathcal{B})$-valued random variable such that $\hat{\beta}$ is conditionally independent of $Y^n$ given $(X,Z)^n$, let $\omega:\mathcal{X}\times \mathcal{B}\to \mathfrak{X}$ be a measurable function and $X_i^{\omega}=\omega(X_i,\hat{\beta})$. Then, we have the following:\\
If $H_0:X\indep Y|Z$ holds, then $H_0^\omega:X^{\omega}\indep Y|Z$ also holds.
\end{theoremE}
\begin{proofE}
    For $X\indep Y|Z$, it follows that $X^n\indep Y^n|Z$. Additionally, for $\hat{\beta}\indep Y^n |(X,Z)^n$, we have by \citet[Lemma~4.3]{dawid1979conditional} that
    \begin{align*}
        (X^n,\hat{\beta})\indep Y^n|Z^n.
    \end{align*}
    Since $\omega$ is a measurable function of $(X_i, \hat{\beta}), i=1,\hdots, n$, we have by \citet[Lemma~4.2]{dawid1979conditional} for $X^{\omega,n}=(\omega(X_i,\hat{\beta}))_{i=1}^n$ that $X^{\omega,n}\indep Y^n|Z^n$, and thus $H_0^\omega$ holds.
\end{proofE}

The general formulation of the theorem via the estimator $\hat{\beta}$ of the parameters of the embedding map allows us to quantify which information we can use to learn the embedding map in addition to the information within $X^n$. In particular, this is more general than for example in \cite{dawid1979conditional}, which allows the function $\omega$ to be only an in-sample function of $X^n$. In contrast, the theorem here also allows it to be, for example, an in-sample function of $(X,Z)^n$. In subsection \ref{subsec:feature_representations}, we explore this further, and translate the theorem to concrete learning approaches for the embedding maps. We now use this theorem to obtain a result on the validity of DNCITs for $\mathcal{P}_0$. Remember that a DNCIT is valid for the null hypothesis $\mathcal{P}_0$ given a level $\alpha\in (0,1)$ and sample size $n$, if
\begin{align*}
    \sup_{\mathbb{P}^S\in \mathcal{P}_0}\mathbb{E}_{\mathbb{P}^S}\left[\mathds{1}\{\varphi_{ \omega, \theta}(S)=1\}\right]=\sup_{\mathbb{P}^S\in \mathcal{P}_0}\mathbb{P}^S(\varphi_{ \omega, \theta}(S)=1)\leq \alpha.
\end{align*}
\begin{theoremE}[Validity of DNCITs][end, no link to proof]\label{thm:validity_DNCITs} 
Let $\hat{\beta}$, $\omega$, and $X^{\omega}$ be as in Theorem \ref{thm:null_hypotheses} and assume that there exists a valid nonparametric CIT $\varphi_\theta^\omega$ for $\mathcal{P}_0^\omega$.
Then, the DNCIT defined  as in \eqref{equ:DNCIT} combining this $\varphi_\theta^\omega$ with $\Bar{\omega}$ is a valid test for $\mathcal{P}_0$.
\end{theoremE}
\begin{proofE} 
Let $\mathbb{P}^{S^\omega|\hat{\beta}}=\mathbb{P}^{S|\hat{\beta}}\circ \Bar{\omega}^{-1}$ be  the conditional probability measure for the measurable map $\Bar{\omega}$ of the sample $S$ onto $S^{\omega(\hat{\beta}, X)}$. Since we obtain from Theorem \ref{thm:null_hypotheses} that $X^{\omega,n}\indep Y^n|Z^n$,
it follows that $\mathbb{P}^{S^\omega|\hat{\beta}}\in \mathcal{P}_0^\omega$. 
Then, 
\begin{align*}
    &\sup_{\mathbb{P}^S\in \mathcal{P}_0}\mathbb{E}_{\mathbb{P}^{S}}\left[\mathds{1}\{\varphi_{ \omega, \theta}\left(S\right)=1\}\right]\\
    =&\sup_{\mathbb{P}^S\in \mathcal{P}_0}\mathbb{E}_{\mathbb{P}^{\hat{\beta}}}\left[\mathbb{E}_{\mathbb{P}^{S|\hat{\beta}}}\left[\mathds{1}\{\varphi_{\omega, \theta}\left(S\right)=1\}\right]\right]\\
    =&\sup_{\mathbb{P}^S\in \mathcal{P}_0}\mathbb{E}_{\mathbb{P}^{\hat{\beta}}}\left[\mathbb{E}_{\mathbb{P}^{S|\hat{\beta}}}[\mathds{1}\{\varphi^\omega_{\theta}\circ \Bar{\omega}(S)=1\}]\right]\\
    =&\sup_{\mathbb{P}^S\in \mathcal{P}_0}\mathbb{E}_{\mathbb{P}^{\hat{\beta}}}\left[\mathbb{E}_{\mathbb{P}^{S|\hat{\beta}}\circ \Bar{\omega}^{-1}}[\mathds{1}\{\varphi^\omega_{\theta}(S^\omega)=1\}]\right]\\
    =&\sup_{\mathbb{P}^S\in \mathcal{P}_0}\mathbb{E}_{\mathbb{P}^{\hat{\beta}}}\left[\underbrace{\mathbb{E}_{\underbrace{\mathbb{P}^{S^\omega|\hat{\beta}}}_{\in\mathcal{P}_0^\omega}}\left[\mathds{1}\left\{\varphi_{ \theta}^\omega\left(S^\omega\right)=1\right\}\right]}_{\leq \alpha \text{ for valid } \varphi_{\theta}^\omega}\right] \leq \alpha
\end{align*}
where we used in line 3 to 4 the integrability of $\varphi^\omega_{\theta}\circ \Bar{\omega}(S)$ under $\mathbb{P}^{S|\hat{\beta}}$ and the resulting equality from \citet[Thm.~3.6.1]{bogachev2007measure}. 
\end{proofE}

This theorem requires the existence of a valid nonparametric CIT for the feature representation and the scalar. Without additional assumptions, there exists no valid nonparametric CIT that has power against any alternative greater than the level of the test \citep{shah2020hardness}. However, making additional assumptions regarding the knowledge of either $\mathbb{P}^{X|Z}$ or $\mathbb{P}^{Y|Z}$ can result in nonparametric CITs that are valid and have greater power than the level against alternatives \citep{candes2018panning}. In the case of DNCITs that assume such additional knowledge, we discuss the availability of this knowledge in our setting in Appendix \ref{app:online}. In addition, we derive in subsection \ref{subsec:nonparam_CITs} the T1E excess of such a DNCIT if an approximation of $\mathbb{P}^{Y|Z}$ can be obtained. For DNCITs based on nonparametric CITs without such additional knowledge, we investigate the corresponding T1E control of the DNCITs in our simulations. Furthermore, we study their application to $S^\omega$ in Appendix \ref{app:online}.

\subsection{Embedding Maps for Deep Nonparametric Conditional Independence Tests}\label{subsec:feature_representations}

The following corollary provides learning approaches for embedding maps that satisfy the assumptions in Theorems \ref{thm:null_hypotheses} and \ref{thm:validity_DNCITs}. Subsequently, we discuss the learning approaches and the impact of such embedding maps on the T1E and power. 

\begin{corollaryE}[][end, no link to proof, text proof={Proof of Corollary \ref{corollary:embeddings}}]\label{corollary:embeddings}
    Let $(\mathcal{A}, \mathcal{F}_\mathcal{A})$ be a measurable space.
    \begin{itemize}
        \item[a)] Let $(\widetilde{X},\widetilde{Y},\widetilde{Z})$ be $(\widetilde{\mathcal{X}},\mathcal{F}_{\widetilde{\mathcal{X}}}), (\widetilde{\mathcal{Y}}, \mathcal{F}_{\widetilde{\mathcal{Y}}}), (\widetilde{\mathcal{Z}},\mathcal{F}_{\widetilde{\mathcal{Z}}})$-valued random variables and assume that $\widetilde{S}=(\widetilde{X}_i,\widetilde{Y}_i,\widetilde{Z}_i)_{i=n+1}^{n+n'}$ is a sample of i.i.d. copies of $(\widetilde{X},\widetilde{Y},\widetilde{Z})$, such that $\widetilde{S}$ is independent of $S$. 
        In particular, $\mathcal{\widetilde{S}}=(\widetilde{\mathcal{X}}, \widetilde{\mathcal{Y}}, \widetilde{\mathcal{Z}})^n$ can be equal to $\mathcal{S}=(\mathcal{X}, \mathcal{Y}, \mathcal{Z})^n$.
        Moreover, let $\hat{\alpha}:\widetilde{\mathcal{S}} \to \mathcal{A}$ be a measurable function. Then, $\hat{\alpha}(\widetilde{S})\indep Y^n|(X,Z)^n$. 
    \item[b)] Let $\hat{\alpha}:(\mathcal{X}\times \mathcal{Z})^n \to \mathcal{A}$ be a measurable function. If $H_0:X\indep Y|Z$ is true, then $\hat{\alpha}((X,Z)^n)\indep Y^n|(X,Z)^n$. 
    \item[c)] Let $\hat{\alpha}:\mathcal{X}^n \to \mathcal{A}$ be a measurable function.  If $H_0:X\indep Y|Z$ is true, then $\hat{\alpha}(X^n)\indep Y^n|(X,Z)^n$. 
    \end{itemize}
    In particular, we obtain for a measurable function $\hat{\beta}:\mathcal{A}\to \mathcal{B}$ that $\hat{\beta}(\hat{\alpha}(\widetilde{S}))\indep Y^n|(X,Z)^n$, $\hat{\beta}(\hat{\alpha}((X,Z)^n))\indep Y^n|(X,Z)^n$ and $\hat{\beta}(\hat{\alpha}(X^n))\indep Y^n|(X,Z)^n$, respectively.
\end{corollaryE}
\begin{proofE}
    \begin{itemize}
        \item[a)] We have $\widetilde{S}\indep S$. Thus, by \citet[lemma~4.2(ii)]{dawid1979conditional}, 
        for the measurable function $U_1:\mathcal{S}\to  (\mathcal{X}\times \mathcal{Z})^n, S\mapsto (X,Z)^n$, we obtain $\widetilde{S}\indep S|(X,Z)^n$.
        Moreover, by \citet[lemma~4.2(i)]{dawid1979conditional}, for the measurable functions $U_2:  \mathcal{S}\to  \mathcal{Y}^n, S\mapsto Y^n$ and $\hat{\alpha}$, 
        it follows:
        \begin{align*}
            \hat{\alpha}(\widetilde{S})\indep Y^n|(X,Z)^n.
        \end{align*}
        \item[b)] Since $X\indep Y|Z$, it follows that $X^n\indep Y^n|Z^n$. Again by \citet[lemma~4.1,4.2]{dawid1979conditional}, we obtain 
        $(X,Z)^n\indep Y^n|(X,Z)^n$. Thus, $\hat{\alpha}((X,Z)^n)\indep Y^n|(X,Z)^n$.
        \item[c)] Since $X\indep Y|Z$, it follows that $X^n\indep Y^n|Z^n$. Thus, $\hat{\alpha}(X^n)\indep Y^n|(X,Z)^n$.
    \end{itemize}
    For a)--c), it follows in particular that $\hat{\beta}(\hat{\alpha}(\cdot))\indep Y^n|(X,Z)^n$, since $\hat{\beta}$ is a measurable function of $\hat{\alpha}$.
\end{proofE}

The corollary allows the use of different learned embedding maps to obtain valid DNCITs. 
Part a) of the corollary allows, for example, the use of sample splitting and transfer learning. 
Parts b) and c) allow, for example, conditional and unconditional unsupervised learning approaches to be applied to the confounder and the image. To be more concrete, $\hat{\alpha}$ could be the estimator of a (conditional) variational autoencoder \citep[cVAE]{kingma2013auto} learned on $(X,Z)^n$, and $\hat{\beta}$ could be the estimator of the parameters of the cVAE's encoder. Then $\omega$ is chosen to map the image to the vector-valued mean in the latent space of the cVAE.
Once feature representations of images obtained by such embedding maps are available, Theorem \ref{thm:validity_DNCITs} ensures the validity of the corresponding DNCITs using valid nonparametric CITs. 
In contrast, when $\hat{\beta}$ is learned from $g(S)$, where $g$ is some measurable function of $S$, it does not hold in general that $X^{\omega,n}\indep Y^n|Z^n$ for such embedding maps. Thus, if the valid nonparametric CITs applied to $S^\omega$ can detect the corresponding induced conditional dependence, the T1E increases and the DNCITs are no longer valid. 

Building on this corollary, we use already available embedding maps learned by sample splitting, transfer, or unsupervised learning, and discuss their benefits and drawbacks within the DNCIT framework. 
In particular, allowing embedding maps that are not DNCIT-specifically learned has several benefits: 
First, it eliminates the need to train DNCIT-specific embedding maps. For example, training embedding maps for 3D structural brain MRI scans can be challenging due to the computational cost, the data acquisition cost, which is about 1000 US dollars per hour of brain MRI data acquisition \citep{marek2022reproducible}, or small sample sizes, as the median sample size of neuroimaging studies is 23 \citep{marek2022reproducible}. The above results allow the translation of embedding maps between data sets, such as the embedding map obtained from the Fastsurfer tool \citep{henschel2020fastsurfer} for brain MRI scans. 
Second, even if we train the embedding maps for a DNCIT, we do not need to split the data for transfer or unsupervised learned embedding maps. This increases the size of the sample $S^\omega$ available for the CIT. 
Third, one may be interested in testing multiple hypotheses between the image $X$ and multiple scalars. Then the feature representation $X^\omega$ can be used repeatedly without the need to train multiple embedding maps for each test. For example, in genome-wide and brain-wide association studies, the goal may be to find causal genes associated with complex phenotypes represented by the image \citep{kirchler2022transfergwas, rakowski2024transfergwas, marek2022reproducible}. The feature representations can then be used to test for conditional associations between the image and each gene or gene region, without having to learn a new embedding map for each test.
Finally, this is consistent with many current practical approaches to medical images using unsupervised and transfer learning \citep{valverde2021transfer, salehi2023study, abrol2021deep, raza2021tour, marek2022reproducible, kirchler2022transfergwas,rakowski2024transfergwas}. Thus, our results provide additional theoretical guarantees and empirical evaluations for these current practices. 

Besides these benefits, we outline two potential pitfalls of not DNCIT-specifically learned embedding maps and address them in the next subsection. One is related to the T1E control, as they often use general embedding maps and then apply a Wald test for the significance of the feature representations in a linear model with $Y$ as response and $X^\omega, Z$ as covariates.  Then, following the proof of theorem \ref{thm:validity_DNCITs}, to control the T1E the tests have to satisfy 
\begin{align*}
    \sup_{\mathbb{P}^S\in \mathcal{P}_0}\mathbb{E}_{\mathbb{P}^{S}}\left[\mathds{1}\{\varphi_{ \omega, \theta}\left(S\right)=1\}\right]
    =\sup_{\mathbb{P}^S\in \mathcal{P}_0}\mathbb{E}_{\mathbb{P}^{\hat{\beta}}}\left[\mathbb{E}_{\underbrace{\mathbb{P}^{S^\omega|\hat{\beta}}}_{\in\mathcal{P}_0^\omega}}\left[\mathds{1}\left\{\varphi_{ \theta}^\omega\left(S^\omega\right)=1\right\}\right]\right],
\end{align*}
i.e. the Wald test has to control the T1E for the distributions $\mathbb{P}^{S^\omega|\hat{\beta}}\in \mathcal{P}_0^\omega$. However, the confounding effects for $X^\omega, Y, Z$ can be highly nonlinear due to the complex, high-dimensional $X$ and the general embedding maps $\omega$. This can lead to falsely rejecting the null hypothesis for tests such as the Wald test which do not take this nonlinear confounding into account and thus, can lead to inflated T1Es.

A second pitfall, which applies more generally to CI testing with embedding maps, is the dependence of the power of the DNCITs on the corresponding embedding maps. 
For any valid CIT testing hypothesis \eqref{test_hypotheses_embedding}, the power increases if elements in $\mathcal{P}_1$ are mapped into regions of $\mathcal{P}_1^{\omega}$ where the corresponding CIT $\varphi_\theta^\omega$ has high power. 
There are a few approaches for two-sample tests that maximize power over embedding maps given the asymptotic distribution of the test statistic of the feature representations under the alternative hypothesis $H_1^\omega$ \citep{liu2020learning}, or via sample splitting or transfer learning explicitly learning an embedding map \citep{kirchler2020two}.
However, this is often not feasible due to unknown test statistic distributions, small sample sizes, or the high computational cost for learning embedding maps, which led to our focus on already learned embedding maps. 
Nevertheless, the choice of such learned embedding maps can lead to a loss in power of the DNCIT if the deviation of CI cannot be detected by the CIT $\varphi_\theta^\omega$. As for the T1E, this is particularly the case due to the potentially strongly nonlinear relationships between $X^\omega$ and $Y$.

Both of these pitfalls in T1E and power are related to a lack of flexibility or too strong (parametric) assumptions, and we therefore address them by applying nonparametric CITs to $S^\omega$ and discussing this application further in subsection \ref{subsec:nonparam_CITs}, section \ref{sec:sim}, section \ref{sec:applications}, and appendix \ref{app:online}.

\subsection{Nonparametric Conditional Independence Tests for Vector-Scalar-Valued Data}\label{subsec:nonparam_CITs}

We apply nonparametric CITs to $S^\omega$ to address the two pitfalls in T1E control and power of DNCITs discussed above. 
In particular, the confounder control of nonparametric CITs typically accounts for strong, nonlinear confounding, thus leading to T1E control over large sets in $\mathcal{P}_0^\omega$ and $\mathcal{P}_0$ for the corresponding DNCITs. 
In addition, the nonparametric CITs can still detect complex conditional dependencies between the feature representations and the scalar in $\mathcal{P}_1^\omega$, which can arise after applying available embedding maps, thus increasing the power of the corresponding DNCITs. 
However, there is no universal best nonparametric CIT, since nonparametric CI testing is a hard problem \citep{shah2020hardness}. 

Our selection of nonparametric CITs for the DNCIT framework is not exhaustive. 
Rather, in order to compare the performance of different categories of CITs within the DNCIT framework, CITs from these categories are selected. 
We selected the RCoT, CPT-KPC, FCIT, and CMIknn, as discussed in subsection \ref{subsec:nonparam_CIT_lit_review}.
For completeness, we discuss in Appendix \ref{app:online}, for each selected CIT, its test statistic and null hypothesis comparison when testing \eqref{test_hypotheses_embedding}, assuming a given fixed embedding map $\omega$, 
to explore their hypothesis restrictions compared to \eqref{test_hypotheses_embedding}, hyperparameter dependencies, time complexity, and the resulting potential advantages and disadvantages of the tests when applied to vector-scalar-valued data, for which the tests are usually not explicitly designed. 
Here, we describe the CPT-based CITs \citep{berrett2020conditional} as an example of nonparametric CITs, and additionally translate the T1E excess result from \citet[Theorem~4]{berrett2020conditional} to the CPT-based DNCITs. 

For the CPT-based CITs, the CPT allows for null comparison of a separately chosen test statistic. The CPT assumes that, in addition to the data $S^\omega$, there exists knowledge on either $\mathbb{P}^{X^{\omega}|Z}$ or $\mathbb{P}^{Y|Z}$ such that one of these distributions is known or can be approximated well. Assuming knowledge of the latter, the idea is to obtain $m=1,\hdots,M$ samples $S^{\omega,m}=(X^{\omega}, Y^{(m)}, Z)^n$ from the null hypothesis $\mathcal{P}_0^\omega$ by permuting  $Y^n$. The distribution of the permutation $\pi$ for $Y^n$ is given by
\begin{align*}
    (Y^{(m)})^n=Y_{\pi^{(m)}}^n, \qquad \mathbb{P}(\pi^{(m)}=\pi|(X,Y,Z)^n)=\frac{p(Y_\pi^n|Z^n)}{\sum_{\pi'\in \mathcal{R}_n} p(Y_{\pi'}^n|Z^n)}
\end{align*}
where $\mathcal{R}_n$ is the set of permutations of $\{1,\hdots,n\}$, $Y_\pi^n=(Y_{\pi(1)},\hdots,Y_{\pi(n)})$ for $\pi\in \mathcal{R}_n$, $p(\cdot|Z^n)\coloneqq p(\cdot|Z_1)\cdots p(\cdot|Z_n)$ and $p(\cdot|Z_i)$ is the density of $\mathbb{P}^{Y_i|Z_i}$, and two efficient sampling algorithms are derived for this distribution \citep[section~4]{berrett2020conditional}.
Then, assuming that we know the true distribution $\mathbb{P}^{Y|Z}$, we obtain a valid CIT for testing \eqref{test_hypotheses_embedding} with p-value as in \citet[thm.~1, 3]{berrett2020conditional} by 
\begin{align}\label{equ:p-value}
    p^\omega\coloneqq \frac{1+\sum_{m=1}^M \mathds{1}\left\{T_\theta^\omega(S^{\omega,m})\geq T_\theta^\omega(S^\omega)\right\}}{1+M}.
\end{align}

Instead of full knowledge of $\mathbb{P}^{Y|Z}$, we assume to have additional observations to estimate $\mathbb{P}^{Y|Z}$ by $\mathbb{\widehat{P}}^{Y|Z}$.
Then, \citet[Theorem~4]{berrett2020conditional} assures for a significance level $\alpha\in [0,1]$ an excess T1E of
\begin{align}\label{equ:CPT_T1E_excess}
    \sup_{\mathbb{P}^{S^\omega}\in \mathcal{P}_0^\omega}\mathbb{P}^{S^{\omega}}(p^\omega\leq \alpha) - \alpha\leq \mathbb{E}_{\mathbb{P}^{(Y,Z)^n}}\left[d_{TV}\left\{\mathbb{{P}}^{Y^n|Z^n}, \mathbb{\widehat{P}}^{Y^n|Z^n}\right\}\right]
\end{align}
where $d_{TV}(\cdot, \cdot)$ denotes the total variation distance between two distributions, $\mathbb{{P}}^{Y^n|Z^n}=\mathbb{{P}}^{Y_1|Z_1}\times\hdots\times\mathbb{{P}}^{Y_n|Z_n}$.

We call the CPT-based DNCITs \texttt{Deep-CPT-}, where we will refer to the chosen test statistic in the part after the second hyphen. For given $\mathbb{P}^{Y|Z}$, Theorem \ref{thm:validity_DNCITs} ensures the validity of the \texttt{Deep-CPT-}s based on the p-values in \eqref{equ:p-value}. Interestingly, we can also translate the result on the excess T1E in \eqref{equ:CPT_T1E_excess} to \texttt{Deep-CPT-}s: 
\begin{theoremE}[T1E bound of the \texttt{Deep-CPT-}][end, no link to proof]\label{thm:deep-cpt_T1E_excess}
    Let $\hat{\beta}$, $\omega$, and $X^{\omega}$ be as in Theorem \ref{thm:null_hypotheses}. Additionally, let $\mathbb{\widehat{P}}^{Y^n|Z^n}$ be an estimate of the true conditional distribution $\mathbb{P}^{Y^n|Z^n}$. Assume that $H_0:X\indep Y|Z$ is true. For the \texttt{Deep-CPT-} and any level $\alpha\in [0,1]$, we obtain
    \begin{align*}
        \mathbb{E}_{\mathbb{P}^S}\left[\mathds{1}\{\varphi_{ \omega, \theta}((X,Y,Z)^n)=1\}\right]\leq \alpha + \mathbb{E}_{\mathbb{P}^{(Y,Z)^n}}\left[d_{TV}\left\{\mathbb{{P}}^{Y^n|Z^n}, \mathbb{\widehat{P}}^{Y^n|Z^n}\right\}\right], \qquad \mathbb{P}^S\in \mathcal{P}_0.
    \end{align*}
\end{theoremE}
\begin{proofE}
    For a chosen level $\alpha$ and CPT-based CITs $\varphi_{ \theta}^\omega$, we can rewrite \eqref{equ:CPT_T1E_excess} to 
    \begin{align*}
        \sup_{\mathbb{P}^{S^\omega}\in \mathcal{P}_0^\omega}\mathbb{E}_{\mathbb{P}^{S^\omega}}\left[\mathds{1}\{\varphi_{ \theta}^\omega(S^\omega)=1\}\right]\leq \alpha +\mathbb{E}_{\mathbb{P}^{(Y,Z)^n}}\left[d_{TV}\left\{\mathbb{{P}}^{Y^n|Z^n}, \mathbb{\widehat{P}}^{Y^n|Z^n}\right\}\right], \qquad \mathbb{P}^{S^\omega}\in \mathcal{P}_0^\omega.
    \end{align*}
    Then, as in the proof of Theorem \ref{thm:validity_DNCITs}, it follows that 
    \begin{align*}
        \sup_{\mathbb{P}^S\in \mathcal{P}_0}\mathbb{E}_{\mathbb{P}^{S}}\left[\mathds{1}\{\varphi_{ \omega, \theta}\left(S\right)=1\}\right]
    \leq \;&\sup_{\mathbb{P}^S\in \mathcal{P}_0}\mathbb{E}_{\mathbb{P}^{\hat{\beta}}}\left[\sup_{\mathbb{\widetilde{P}}\in\mathcal{P}_0^\omega}\mathbb{E}_{\mathbb{\widetilde{P}}}\left[\mathds{1}\{\varphi_{ \theta}^\omega(S^\omega)=1\}\right]\right]\\
    \leq\; &\sup_{\mathbb{P}^S\in \mathcal{P}_0}\mathbb{E}_{\mathbb{P}^{\hat{\beta}}}\left[\alpha +\mathbb{E}_{\mathbb{P}^{(Y,Z)^n}}\left[d_{TV}\left\{\mathbb{{P}}^{Y^n|Z^n}, \mathbb{\widehat{P}}^{Y^n|Z^n}\right\}\right]\right]\\
    =\; &\alpha +\mathbb{E}_{\mathbb{P}^{(Y,Z)^n}}\left[d_{TV}\left\{\mathbb{{P}}^{Y^n|Z^n}, \mathbb{\widehat{P}}^{Y^n|Z^n}\right\}\right].
    \end{align*}
\end{proofE}

This result on the excess T1E explicitly shows that \texttt{Deep-CPT-}s targeting $\mathbb{P}^{Y|Z}$
simplify the null comparison significantly, since the complexity of $X$ and $X^\omega$ does not impact the T1E control, which can be an advantage over other CITs which additionally consider $X^\omega$ in the null comparison. 
In combination with the CPT, the remaining choice is that of a suitable test statistic, typically a measure of conditional dependence between $X^\omega$ and $Y$ given $Z$, which we discuss together with additional details on the CPT in Appendix \ref{app:cpt-kpc}. 

\section{Simulation Study}\label{sec:sim}

We examine the empirical performance of the DNCITs on simulated data. The DNCIT R-package is available at \url{https://github.com/MSimnach/DNCIT} and the code to reproduce all the results in this and the next section can be found at \url{https://github.com/MSimnach/dncitPaper}. Throughout our empirical studies we use data from the UKB \citep{sudlow2015uk}, and for the images particularly the brain imaging data contained therein. There are many detailed descriptions of the UKB brain imaging data \citep[e.g.]{miller2016multimodal}, and we rely on standardized brain imaging pipelines \citep{henschel2020fastsurfer, fischl2012freesurfer}, which were also developed and executed on behalf of the UKB \citep{alfaro2018image, alfaro2021confound, smith_ukb_description}.

\subsection{Design}

We compare the proposed DNCITs to each other and to a commonly used baseline on brain imaging data from the UKB. We follow the recommendations and terminology for simulation studies in biostatistics in \citet{morris2019using}. Further details on specific design choices can be found in the Appendix \ref{app:sim_realism}.

\textbf{Aims:} 
To evaluate the performance of the DNCITs I) for different nonparametric CITs  and II) for different embedding maps in realistic DGMs with respect to A) increasingly complex confounding effects between $Z$ and $Y$; 
and B) increasing confounder dimension.

\textbf{Data generating mechanisms (DGMs):} We consider 18 DGMs. For all, we vary the sample size roughly logarithmically over $n=$ 
145, 
256, 350, 460,
825, 1100, 1475, 1964, 5000, 10000. 
These DGMs extend simulation studies to evaluate CITs for low-dimensional $Y$ and $X$ \citep{bellot2019conditional, strobl2019approximate, shah2020hardness} to high-dimensional $X$.

For the images $X$ and confounders $Z$, we repeatedly resample with replacement from the UKB to obtain realistic images and correlation structures with the confounders. In the following, individual realisations will be denoted with lowercase letters. As images $\bm{X}=(\bm{x}_1^\top,\hdots,\bm{x}_n^\top)^\top$, we resample T1-weighted structural brain MRI scans such that $\bm{x}_i\in \mathbb{R}^{256\times 256\times 256}$. Furthermore, the confounders $\mathbf{Z}=(\mathbf{z}_1^\top,\hdots,\mathbf{z}_n^\top)^\top$ are resampled as subsets of age, head size, sex, assessment site, assessment date, MRI quality control metric, head location in scanner (4 features), and the first five genetic PCs, s.t. $\mathbf{z}_i\in \mathbb{R}^k, k=1,2,4,6,10,15$ for six different settings with increasing confounder dimension. These confounders are chosen to make the results relevant to a broader literature,  because they are commonly selected confounders for brain imaging data \citep{hyatt2020quandary, alfaro2021confound}, they are among the relevant confounders identified in the literature related to our behavioral traits and confounder control applications \citep{avinun2020little, alfaro2021confound}, and they are commonly adjusted for in alternative potential applications such as GWAS \citep{kirchler2022transfergwas, rakowski2024transfergwas}. 

Then, we simulate $\mathbf{
y}=(y_1,\hdots,y_n)^\top$ from the confounders $\mathbf{Z}$  and feature representations $\mathbf{X}^{\widetilde{\omega}}=(\mathbf{x}^{\widetilde{\omega},\top}_1,\hdots,\mathbf{x}^{\widetilde{\omega},\top}_n)^\top,$ $\mathbf{x}^{\widetilde{\omega}}_i\in \mathbb{R}^{139}$ of $\bm{X}$, where $\widetilde{\omega}$ is the Fastsurfer embedding map \citep{henschel2020fastsurfer} which maps the images onto feature representations of size $q=139$. In particular, $\mathbf{y}$ is simulated for $c=0$ (CI) and $c=1$ (no CI) as
\begin{align}\label{equ:sim_setting}
\begin{aligned}
y_i&=c\mathbf{x}^{\widetilde{\omega},\top}_i\mathbf{w}_x+g_z(\mathbf{z}_i)\mathbf{w}_z+\varepsilon_i\\
&\text{a) linear: } g_z(\mathbf{s})= \mathbf{s}^\top,\qquad \text{b) squared: } g_z(\mathbf{s})=(\mathbf{s}^\top,(s_j^2)_{j\in\mathcal{J}_c}) , \\
&\text{c) complex: } g_z(\mathbf{s})=(\mathbf{s}^\top,(s_j^2, s_{j}s_{sex})_{j\in\mathcal{J}_c},s_{date}^3,s_{date}^4)\\
\mathbf{w}_x&=(w_{x,1},\hdots,w_{x,q}),\; w_{x,j}=a_{x,j}\frac{|\delta_{x,j}|}{\sum_{j=1}^{q}|\delta_{x,j}|}, \;\delta_x\sim N(\mathbf{0}, \mathbf{I}_{q}),\;  a_{x,j}\sim Ber(0.5) \\
\mathbf{w}_z&=(w_{z,1},\hdots,w_{z,p_z}),\; w_{z.j}=\frac{|\delta_{z,j}|}{\sum_{j=1}^{p_z}|\delta_{z,j}|}, \;\delta_z\sim N(\mathbf{0}, \mathbf{I}_{p_z}) 
\end{aligned}
\end{align}
where $p_z=\dim{(g_z(\mathbf{s}))}$, $\mathcal{J}_c$ denotes the index set of continuous variables, and we standardize $\mathbf{X}^{\widetilde{\omega}}, \mathbf{Z}$ column-wise beforehand.
The standardizations as well as the normalizations in $\mathbf{w}_x$ and $\mathbf{w}_z $ balance the effect sizes on $\mathbf{y}$ within $\mathbf{X}^{\widetilde{\omega}}$ and between $\mathbf{X}^{\widetilde{\omega}}$ and $\mathbf{Z}$ across the DGMs. 

To address aim A), we increase the complexity of the confounding effects of $Z$ on $Y$ via the functions $g_z$ for the six confounders age, head size, sex, assessment site, assessment date, and MRI quality control metric. For the confounding effects $g_z(\mathbf{s})=(\mathbf{s}^\top,(s_j^2)_{j\in\mathcal{J}_c})$, we increase the confounder dimension over $1,2,4,6,10,15$ to evaluate aim B). For both variations, we consider CI ($c=0$) and no CI ($c=1$).

\textbf{DNCITs:} We vary the feature representations $\mathbf{X}^{\omega}$ over four different embedding maps $\omega$, since $\mathbf{X}^{\widetilde{\omega}}$, which is used to simulate conditional dependence, is typically unavailable in practice.
First, we set $\mathbf{X}^\omega=\mathbf{X}^{\widetilde{\omega}}$ to establish a baseline and to evaluate nonparametric CITs when applied to feature representations, for which they are originally not designed.
Second, we add random noise to the feature representation used in the nonparametric CITs to mimic noisy estimation of the embedding map by setting $\mathbf{X}^\omega=\mathbf{X}^{\widetilde{\omega}}+\varepsilon_\mathbf{x}, \varepsilon_\mathbf{x}\sim N(0,\sigma_{\varepsilon_\mathbf{x}}^2I_q), \sigma_{\varepsilon_\mathbf{x}}^2=3$.  
Third, we use parts of the Freesurfer embedding map \citep{fischl2012freesurfer, alfaro2018image} as feature representation $\mathbf{X}^\omega\in \mathbb{R}^{n\times 165}$. 
Finally, the feature representation $\mathbf{X}^\omega\in \mathbb{R}^{n\times 256}$ is obtained from the latent representation of a cVAE trained on the ADNI dataset \citep{mueller2005alzheimer} for the work of \citet{rakowski2024transfergwas}.

In addition, we vary the nonparametric CITs as discussed in subsection \ref{subsec:nonparam_CIT_lit_review} and \ref{subsec:nonparam_CITs} with implementation details given in Appendix \ref{app:online}. Thus, we apply the DNCITs \texttt{Deep-RCoT, Deep-CPT-KPC, Deep-CMIknn} and  \texttt{Deep-FCIT}, where for a given embedding map, in the name we replace Deep with the corresponding embedding map used in the DNCIT. Additionally, the DNCIT \texttt{Deep-WALD} with a parametric Wald test for the significance of all features in the feature representation is used as a baseline. 

\textbf{Target and performance measures:} In our simulation study, we evaluate all DNCITs w.r.t. their performance testing the hypothesis in \eqref{test_hypotheses_image}. Additionally, we list the hypotheses in \eqref{test_hypotheses_embedding} the different nonparametric CITs target in practice in appendix \ref{app:online}. 

Performance is measured in terms of the T1E, power and runtime. T1E and power are estimated from the rejection rates 
\begin{align*}
    \widehat{RR}=\frac{1}{n_{\text{sim}}}\sum_{i=1}^{n_{\text{sim}}}\mathds{1}\{p_i\leq \alpha\}
\end{align*}
under the null ($c=0$) and alternative ($c=1$) hypothesis, respectively, for a significance level of $\alpha=0.05$, over $i=1,\hdots,n_{\text{sim}}$ random generations and p-values $p_i$ for each DGM. We choose $n_{\text{sim}}=200$ for computational reasons, resulting in a maximum Monte Carlo standard error \citep[sec.~5.3]{morris2019using} of
\begin{align*}
    \text{Monte Carlo SE}(\widehat{RR})=\sqrt{\frac{\widehat{RR}(1-\widehat{RR})}{n_{\text{sim}}}}\leq \sqrt{\frac{0.5^2}{200}}\approx 0.035
\end{align*}
for an estimated rejection rate of 0.5.
Finally, runtime is measured in seconds on a high-performance server equipped with a 112 core Intel(R) Xeon(R) Gold 6330 CPU @ 2.00GHz and 502 GB of RAM.

\subsection{Results}

\begin{figure}
    \centering
    \includegraphics[width=\linewidth]{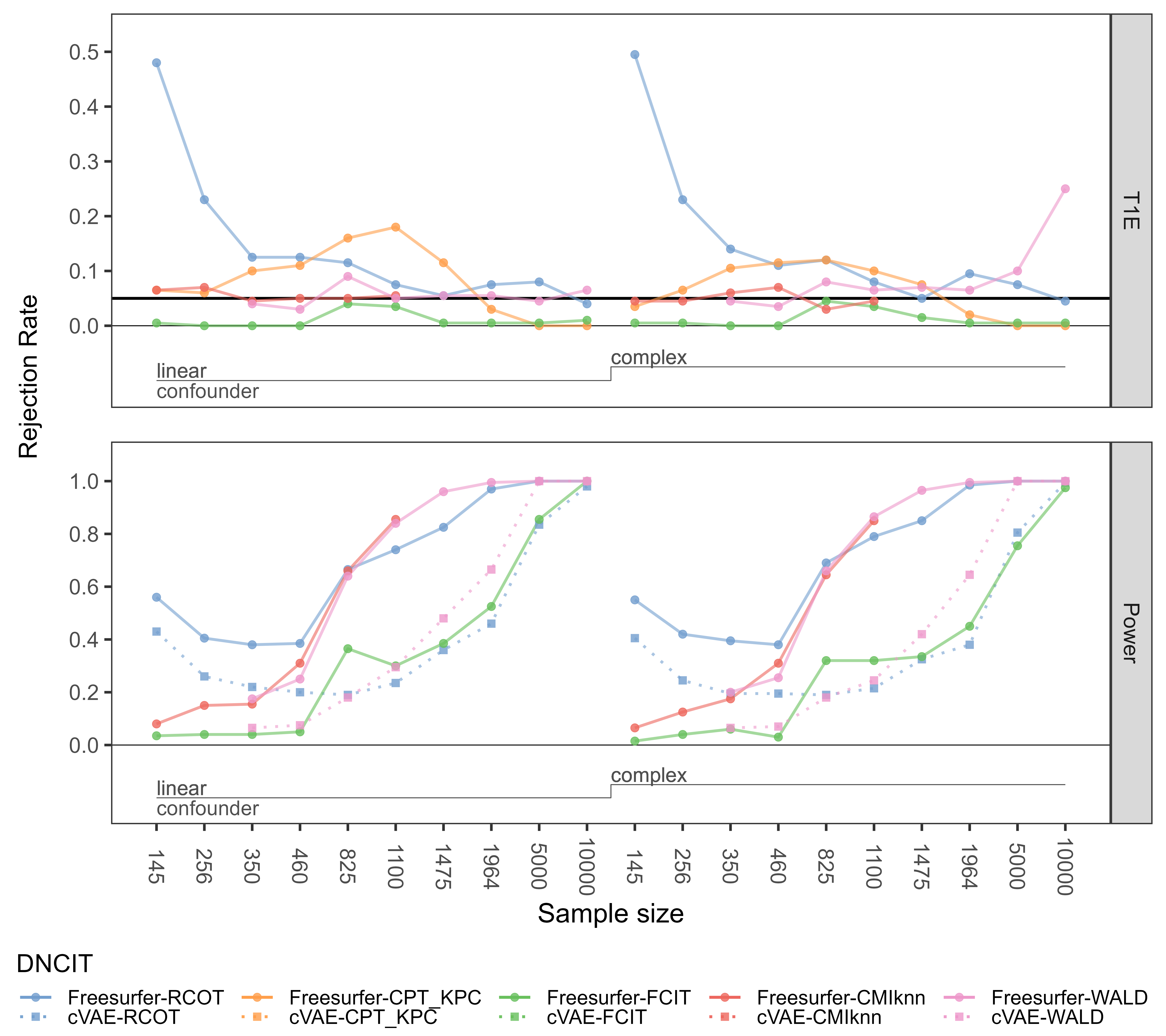}
    \caption{The rejection rates of the DNCITs for $c=0$ (CI, top) and $c=1$ (no CI, bottom) for increasingly complex confounder relationships (columns). For each column, the sample size is increased from left to right. The confounder dimension is set to 6. Horizontal lines at 0 and $\alpha=0.05$.}
    \label{fig:sim_conf_relation}
\end{figure}

\begin{figure}
    \centering
    \includegraphics[width=\linewidth]{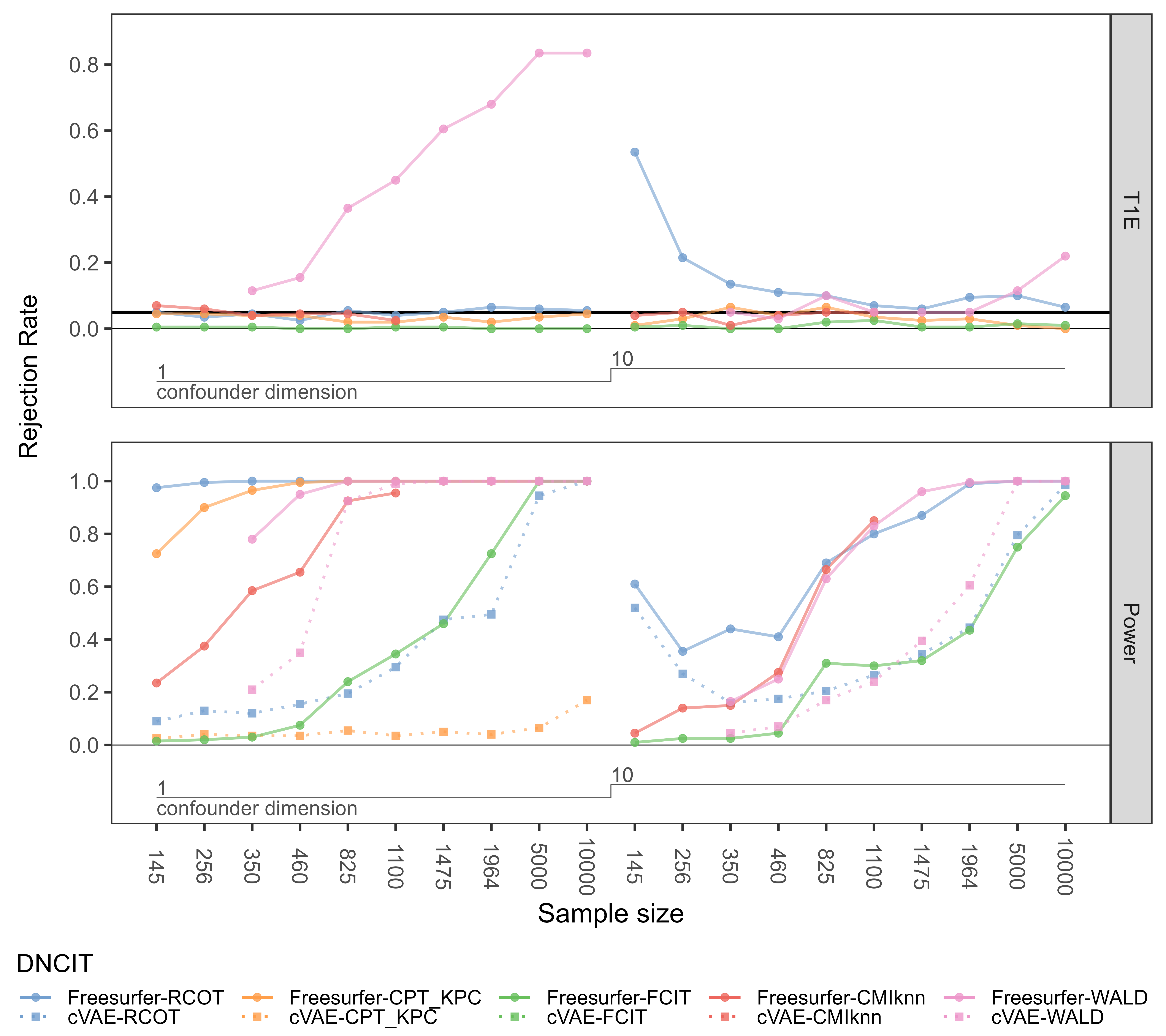}
    \caption{The rejection rates of the DNCITs for $c=0$ (CI, top) and $c=1$ (no CI, bottom) for increasing confounder dimension (columns). For each column, the sample size is increased from left to right. The confounder relationship is set to $g_z(\mathbf{s})=(\mathbf{s}^\top,(s_j^2)_{j\in\mathcal{J}_c})$, where $\mathcal{J}_c$ denotes the index set of continuous variables. Horizontal lines at 0 and $\alpha=0.05$.}
    \label{fig:sim_conf_dim}
\end{figure}

\begin{figure}
    \centering
    \includegraphics[width=0.65\linewidth]{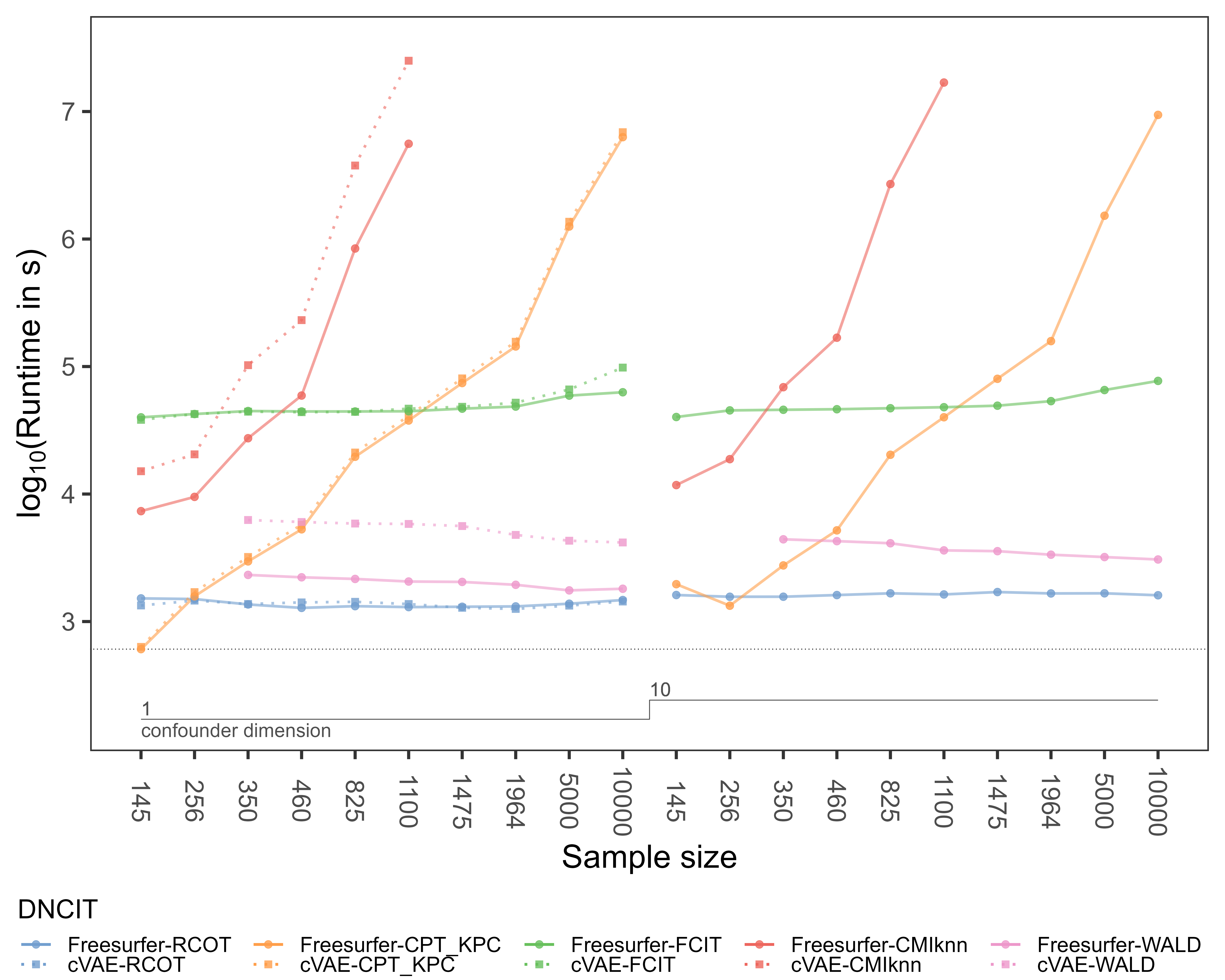}
    \caption{The logarithm of the runtime of the DNCITs for $c=0$ (CI) for increasing confounder dimension (columns). Each column increases the sample size from left to right. The confounder relationship is set to $g_z(\mathbf{s})=(\mathbf{s}^\top,(s_j^2)_{j\in\mathcal{J}_c})$, where $\mathcal{J}_c$ denotes the index set of continuous variables.}
    \label{fig:sim_conf_dim_runtime}
\end{figure}

First, note that the DNCITs differ in the sample sizes to which they can be applied in their current implementation. The \texttt{Deep-Wald} is not applicable to sample sizes less than or equal to the dimension of the embedding map, i.e. here 256 for the cVAE embedding map, plus the dimension of the confounder. The \texttt{Deep-CMIknn} based on a knn search is not practically feasible for sample sizes larger than 1100 due to computational constraints, where we used the runtime of the \texttt{Deep-CMIknn} of about 37 minutes per run for sample size 1100 as a cutoff, also shown in Figure \ref{fig:sim_conf_dim_runtime}. The \texttt{Deep-FCIT} can throw errors for sample sizes up to 460 due to the implementation of the prediction model, especially in the case of CI, see also the Figure \ref{fig:qq-plot_dim6_squared_freesurfer} and \ref{fig:qq-plot_dim6_squared_freesurfer_no_ci}. Finally, the \texttt{Deep-RCoT} and \texttt{Deep-CPT-KPC} are applicable to all sample sizes. Second, as the results for the Fastsurfer and noisy Fastsurfer embedding maps closely resemble those of the Freesurfer embedding map, they are not presented here. Nevertheless, this suggests that the DNCITs are not sensitive to noise or slight variations in embedding maps compared to the true embedding map. Third, we only display DNCITs with the Freesurfer embedding map for T1E, as there was no significant variation in T1E across the different embedding maps. Finally, to enhance the clarity of the figures, we exclude DNCITs with a power less than 0.15 for the largest sample size in the power plots. Accordingly, we do not show the \texttt{cVAE-CPT-KPC}, \texttt{cVAE-FCIT} and \texttt{cVAE-CMIknn} in the Figures \ref{fig:sim_conf_relation} and \ref{fig:sim_conf_dim}, nor the \texttt{Freesurfer-CPT-KPC}, with the exception of confounder dimension 1 in Figure \ref{fig:sim_conf_dim}. We present more detailed results including QQ-plots of the p-values in Appendix \ref{app:detailed_sim}.

We first present the \textbf{general results} for embedding maps (represented by different line types and point shapes) and nonparametric CITs (represented by different colors), focusing on the fixed complex confounder relationship $g_z(\mathbf{s})=(\mathbf{s}^\top,(s_j^2, s_{j}s_{sex})_{j\in\mathcal{J}_c},s_{date}^3,s_{date}^4)$ and a confounder dimension of 6 (right column in Figure \ref{fig:sim_conf_relation}). The \texttt{Deep-RCoT} (blue) shows inflated T1Es for smaller sample sizes (up to 1100), likely due to its reliance on asymptotic approximations. In addition, it shows slightly inflated T1Es for sample sizes 1964 and 5000 when confounder dimensions are larger than two, possibly due to the fixed confounder kernel embedding dimension across varying confounder dimensions. Adjusting this kernel embedding dimension to scale with the confounder dimension could improve T1E control, as discussed in appendix \ref{app:rcot} and \citet[sec.~7]{strobl2019approximate}. In contrast, the \texttt{Deep-Wald} (pink) exhibits inflated T1Es for larger sample sizes (over 1475) because we only adjust for confounders linearly in the (parametric) \texttt{Deep-Wald}. Despite these limitations, both tests have comparable power and are the most powerful among all tests for the sample sizes for which they have no or only slightly inflated T1Es. As expected, the power decreases as the association between the true Fastsurfer embedding map and the chosen embedding map for testing weakens, e.g.\ the power for  the unrelated cVAE (square, dashed) is smaller than for  the more closely related Freesurfer map (point, solid). However, even for the cVAE embedding map, these two tests approach power one at the larger sample sizes. The \texttt{Deep-FCIT} (green) controls the T1E but has low power across all embedding maps. The \texttt{Deep-CPT-KPC} (orange) shows inflated T1E for medium sample sizes, potentially because the KPC is measuring conditional associations that the CPT does not control for accurately at these sample sizes, compare also appendix \ref{app:cpt-kpc}. Lastly, the \texttt{Deep-CMIknn} (red) maintains T1E control for all embedding maps and shows power comparable to \texttt{Deep-RCoT} and \texttt{Deep-Wald} for the Freesurfer map. However, for the cVAE, its power approaches the significance level, indicating that the \texttt{Deep-CMIknn} is more sensitive to the similarity of the chosen and true embedding maps.

The results for the squared and complex \textbf{confounder relationship} are consistent, so we focus on the linear and complex relationship, as shown in Figure \ref{fig:sim_conf_relation}.
Across all embedding maps, the confounder relationship has a similar effect on the performance of the DNCITs. 
Additionally, the power of all nonparametric CITs remains stable across different confounder relationships. 
The \texttt{Deep-CMIknn} and \texttt{Deep-FCIT} effectively control T1Es for the confounder relationships considered, although the latter is overly conservative with rejection rates close to zero.
The inflation of T1Es for the \texttt{Deep-CPT-KPC} is slightly higher for linear confounding effects, possibly because the KPC is better at detecting them, see also appendix \ref{app:cpt-kpc}.
The \texttt{Deep-RCoT} shows inflated T1Es for smaller sample sizes (up to 1100) across all confounder relationships. 
In addition, the \texttt{Deep-Wald} shows inflated T1Es for larger sample sizes (over 1475) in cases of nonlinear confounding.

The results for \textbf{confounder dimensions} 1 and 2 as well as 4, 6, 10, and 15, respectively, are consistent, so we focus on dimensions 1, and 10, as shown in Figure \ref{fig:sim_conf_dim}.
Across all embedding maps, increasing the confounder dimension has similar effects on performance.
The \texttt{Deep-RCoT} performs worse with increasing confounder dimensions, especially with an inflation of T1Es for small sample sizes. As previously stated in the paragraph on general results, this behavior is to be expected and could potentially be improved by adjusting the dimension of the kernel embedding of the confounder, compare appendix \ref{app:rcot} and \citet[sec.~7]{strobl2019approximate}.
The \texttt{Deep-CPT-KPC} controls T1Es across all dimensions but is highly sensitive to the confounder dimension in terms of power. While it performs similarly to \texttt{Deep-RCoT} and \texttt{Deep-WALD} for a one-dimensional confounder using the Freesurfer embedding maps, it loses almost all power as the confounder dimension increases. Since its power relies on the KPC test statistic, it indicates that the KPC struggles with detecting conditional associations for higher-dimensional confounders and one-dimensional $Y$, potentially due to the limitations of the knn search in $\mathcal{Y} \times \mathcal{Z}$.
The \texttt{Deep-CMIknn} controls T1Es for all confounder dimensions, with power varying as the confounder dimension grows. This could be attributed to the knn matching in $\mathcal{Z}$, reflecting a trade-off between better T1E control for smaller $k$ and larger power for larger $k$ as the closeness of nearest neighbors varies \citep{runge2018conditional}.
The \texttt{Deep-FCIT} exhibitis consistent T1E control and power across all dimensions. 
Lastly, the \texttt{Deep-WALD} shows inflated T1Es, with the inflation worsening as the sample size grows. This issue is particularly evident in cases with lower confounder dimensions, since for confounder dimensions of one and two all confounders are continuous, and thus the linearity assumption of the Wald test in the confounders is violated due to the squared confounder relationship $g_z(\mathbf{s}) = (\mathbf{s}^\top, (s_j^2)_{j\in \mathcal{J}_c})$. The power of the \texttt{Deep-WALD} test remains stable and relatively high compared to the other tests across all confounder dimensions.

Figure \ref{fig:sim_conf_dim_runtime} shows the \textbf{runtime} of all DNCITs.
The results for \textbf{confounder dimensions} 2, 4, 6, 10, and 15 are consistent, so we focus on dimensions 1, and 10.
The embedding maps and confounder relationships have a minimal effect on runtime, so the results are presented for the cVAE and Freesurfer embedding map for confounder dimension 1, and for dimension 10 using only the Freesurfer embedding map, and the confounder relationship is set to $g_z(\mathbf{s})=(\mathbf{s}^\top,(s_j^2)_{j\in\mathcal{J}_c})$.
Across all embedding maps and confounder dimensions, the runtime increases significantly with larger sample sizes for \texttt{Deep-CMIknn} and \texttt{Deep-CPT-KPC}, both of which use knn algorithms that scale quadratically with sample size. 
The \texttt{Deep-FCIT} also has a relatively long runtime, which further increases for sample sizes above 1100. 
The \texttt{Deep-Wald} exhibits the most notable runtime growth with increasing confounder and feature representation dimensions, but constant runtime across sample sizes. 
Finally, the \texttt{Deep-RCoT} maintains consistently low runtimes across all sample sizes and confounder dimensions compared to the other tests.

\subsection{Conclusion:} 
For different embedding maps, the DNCITs are affected similarly by confounder relationship and dimension. Varying the embedding maps has only minor effects on the T1Es, but the power of the DNCITs can vary significantly with changes in the embedding map, and the size of this effect depends on the specific nonparametric CIT used.

The performance of the DNCITs is highly dependent on the chosen CIT, with each CIT responding differently to confounder relationships and dimensions. In this study, \texttt{Deep-RCoT} performed best for sample sizes above 1100 and is computationally efficient, while the computationally more expensive \texttt{Deep-CMIknn} was more effective for smaller sample sizes. 
In contrast, the \texttt{Deep-CPT-KPC} and \texttt{Deep-FCIT} lack power in almost all settings. 
The \texttt{Deep-Wald}, often used to test for conditional associations while controlling for confounders, has strongly inflated T1Es when parametric assumptions are not fulfilled, and is thus not generally recommended.  

\section{Real-World Applications}\label{sec:applications}

We apply the DNCITs in two applications to evaluate their empirical performance on real-world datasets.

\subsection{Brain Structure and Behavioral Traits}\label{subsec:brain_behavior}

This subsection applies DNCITs to study significant links between brain structure and behavioral traits for healthy individuals in the UKB. Such links are important in the field of personality neuroscience and rely on the idea of the brain as the proximal source of behavior; compare, for example, \citet{yarkoni2015neurobiological}. 

There exist several studies showing associations between behavioral traits such as neuroticism and brain structure \citep{young2022personality, zhang2023personality}. However, these findings could not be replicated in healthy individuals from larger cohorts in systematic replication studies \citep{kharabian2019empirical, genon2022linking}. In particular, little evidence was found for associations between brain structures and the big five behavioral traits \citep[BFBTs; ][]{digman1990personality, avinun2020little}. 
According to \citet{genon2022linking}, this ambiguity may arise because finding replicable effects in studies with small sample sizes is often unrealistic due to insufficient power of existing statistical tests. Furthermore, several studies have focused on one-to-one mappings between a specific brain structure and a psychometric measure such as a behavioral trait. 
This ignores potential interactions between brain structures by relying on linear, often univariate statistical methods, resulting in lower power for the test on the joint significance of all brain structures and the behavioral traits, in particular when also adjusting for multiple testing.

To address these shortcomings of existing studies, we use the larger cohort of the UKB, increasing the sample size for testing from 1107 in the largest replication study \citep{avinun2020little} to 8634. Additionally, as shown in the simulation study, DNCITs may have greater power to detect nonlinear, distributional relationships between brain structures and BFBTs compared to linear \citep{avinun2020little} or machine learning prediction-based CITs similar to the FCIT \citep{genon2022linking}. Finally, DNCITs allow us to test for conditional associations between the whole brain MRI scan and each BFBT, thereby maximizing the potential effect sizes by including nonlinear effects and interactions within the whole scan. 

In the following, we first replicate the study in \citet{avinun2020little}, which found little evidence for associations between certain brain structures derived from brain MRI scans and each BFBT on a cohort of healthy university students obtained from the Duke Neurogenetics Study, on the larger cohort of healthy subjects from the UKB. 
We then extend the study to associations between the whole MRI scans and the BFBTs applying the \texttt{Freesurfer-RCoT} and \texttt{Fastsurfer-RCoT}, which showed T1E control and the highest power for sample sizes as large as in the UK Biobank.

\subsubsection{Brain morphometry, Behavioral Proxies and Confounders}\label{subsec:brain_behaviour_data_descrip}

To obtain a sample of healthy subjects, we follow the preprocessing in \citet[sec.~2.1]{avinun2020little}. 
Thus, we excluded participants with diagnoses of cancer, stroke, diabetes requiring insulin, chronic kidney or liver disease, those taking psychotropic or glucocorticoid medications, and those with personality and psychiatric disorders based on the information from \citet[Supplemental Table 3]{sekimitsu2022interaction}, \citet{meulen2022association}, and \citet[Supplemental Table 1]{ruijter2022association}.

Using brain morphometric measures from the UKB that are analogous to those in \citet{avinun2020little}, we selected 107 measures, including surface area, cortical thickness, subcortical volume, and white matter microstructural integrity, based on a subset of the Freesurfer embedding map. We excluded individuals lacking these brain measures, resulting in a sample size of 8634 to replicate the study by \citet{avinun2020little}. For the DNCIT analysis, we used the feature representations derived from the Fastsurfer embedding map as a robustness check against the Freesurfer embedding map. After excluding individuals without these feature representations, we obtained a sample of 9115, as the Fastsurfer embedding map was available for a slightly larger number of individuals in the UKB.

To obtain BFBT scores, we used a neuroticism score (0-12) from the UKB touchscreen questionnaire during recruitment \citep{smith2013prevalence}. Additionally, we created BFBT proxies for sociability, diligence, and curiosity (0-4), and warmth and nervousness (0-5), similar to those in \citet{john1999big}. We followed \citet{dahlen2022influence, ruijter2022association}, who used these proxies to study myocardial infarction and stroke risks in the UKB cohort. This results in six BFBT proxies used throughout the analysis.

To include confounders, our study followed the confounding instructions and the conventional set of confounders for the Wald test for brain MRI scans of the UKB \citep[sec.~2.9]{alfaro2021confound}. Specifically, we included age, sex, assessment date, assessment center, head size, head location measures, and MRI scan quality control discrepancy as confounders. Additionally, squared age, squared assessment date, and an interaction term between age and sex were included for all Wald tests. We also included ethnicity information using the first five genetic PCs of whole-genome SNPs derived in the UKB, similar to \citet{avinun2020little}, who used the first four multidimensional scaling coefficients.

\subsubsection{Analysis}

To replicate \citet{avinun2020little}, we used a Wald test to assess the significance of each of the 107 brain morphometric measures when each of the BFBTs was regressed on each measure along with the confounders. Additionally, we used the \texttt{Freesurfer-WALD} DNCIT to test for the joint significance of all brain measures when each BFBT was regressed on all brain measures together with the confounders, thereby accounting for linear additive effects of the confounder set, which includes also squared terms for age and the assessment date and an interaction term between age and sex. To compare our new DNCITs, we applied the \texttt{Freesurfer-RCoT} to all brain measures and each BFBT, given the confounders, to capture nonlinear interactions between the brain measures as well as nonlinear relationships between brain measures and each BFBT. A robustness analysis was conducted using the \texttt{Fastsurfer-RCoT} to test for conditional associations between the brain MRI scans and each BFBT, as well as collectively across all BFBTs by applying the \texttt{Freesurfer-RCoT} 
 and \texttt{Fastsurfer-RCoT} to the brain MRI scans, all BFBTs, and the confounders. The latter is directly possible with the \texttt{Deep-RCoT}, see Appendix \ref{app:rcot} for details. We selected the RCoT as the nonparametric CIT throughout this subsection since it performed best in the simulation study for sample sizes greater than 1100.
 
\subsubsection{Results}

\begin{figure}
 \centering
    \begin{subfigure}[t]{0.49\textwidth}
        \centering
        \includegraphics[width=\textwidth]{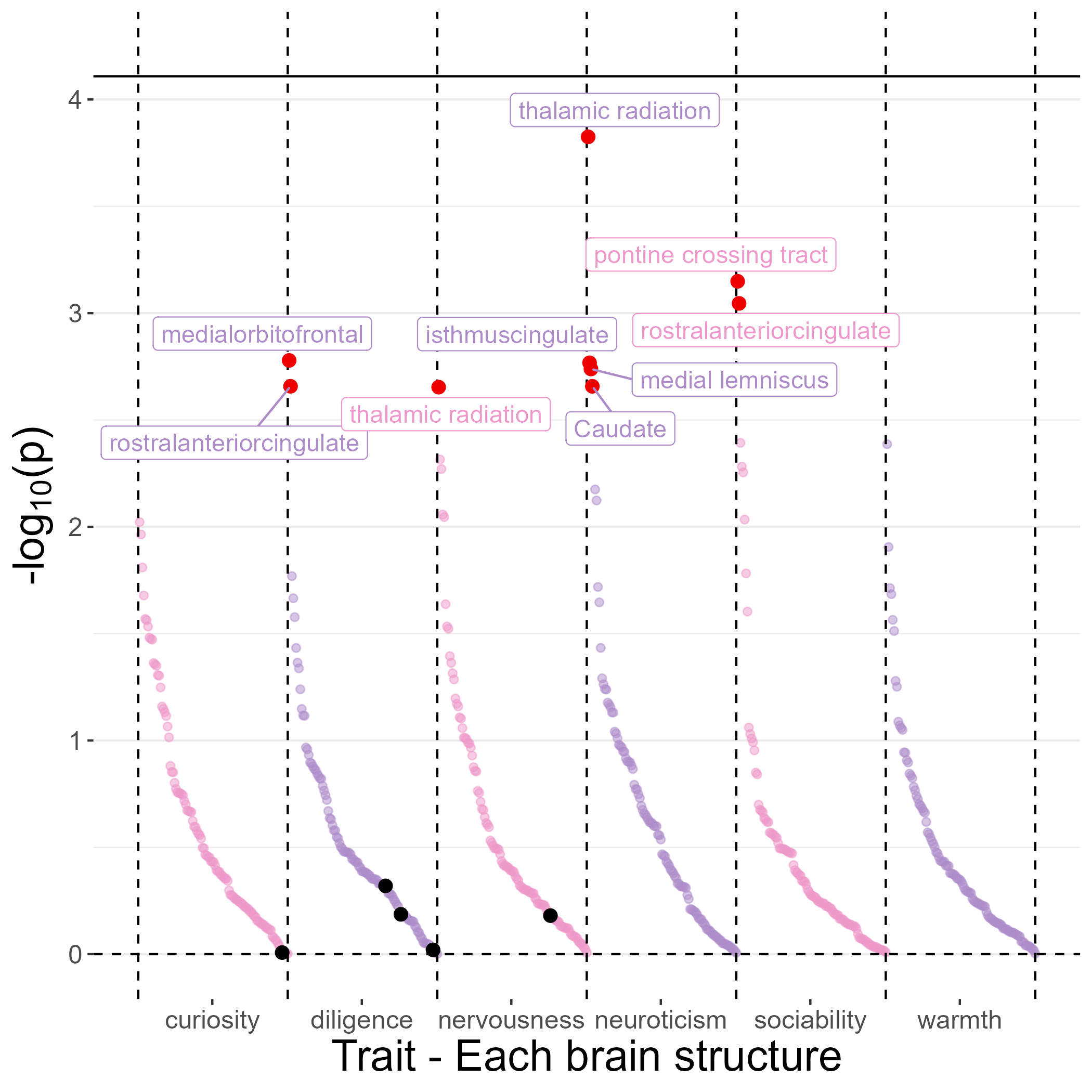}
    \end{subfigure}
    \hfill 
    \begin{subfigure}[t]{0.49\textwidth}
        \centering
        \includegraphics[width=\textwidth]{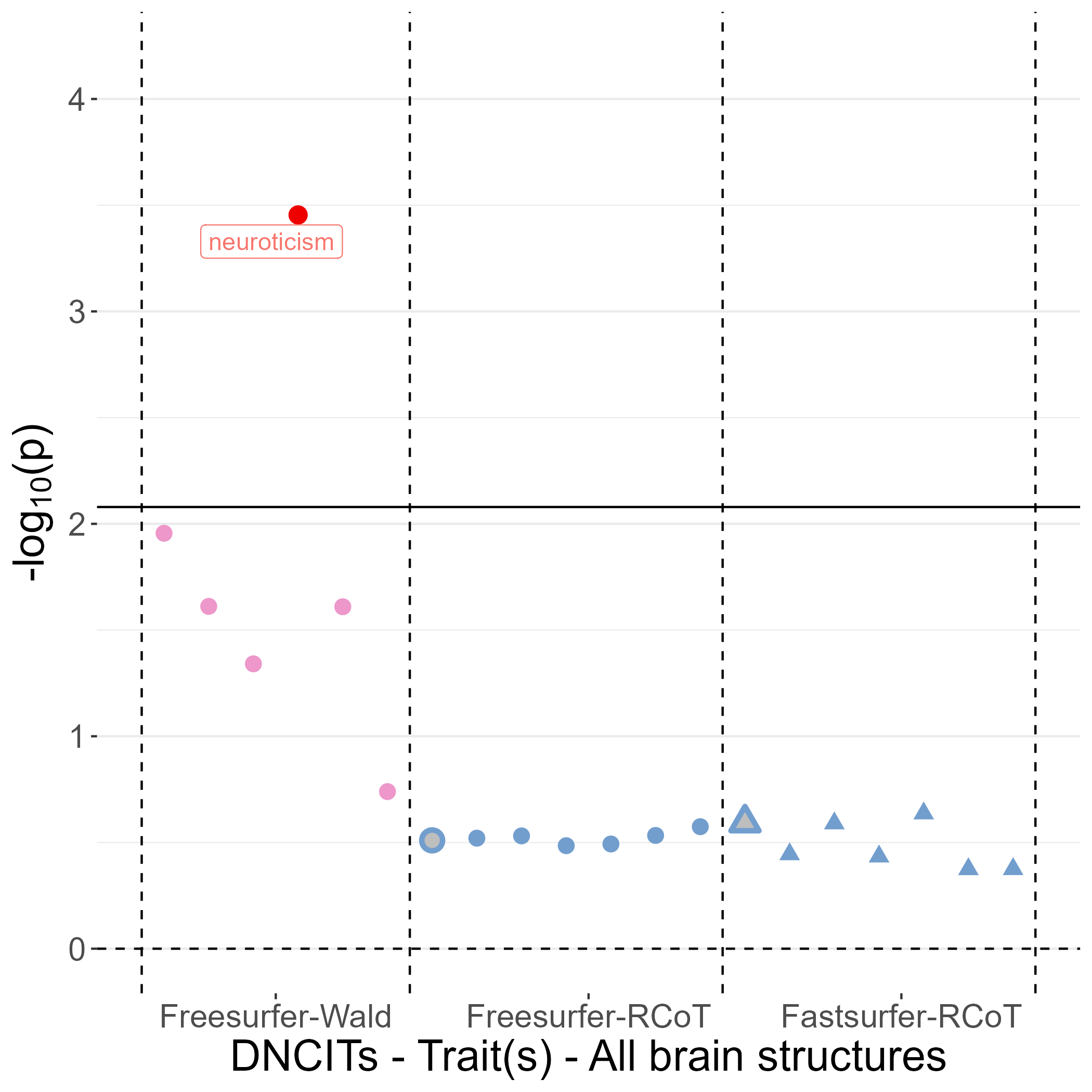}
    \end{subfigure}

\caption{In the left figure, $-\log_{10}(p)$ values of individual Wald tests are shown for each trait-brain structure combination, sorted by size for each trait. The trait-structure combinations identified as significant in \citet{avinun2020little} are highlighted in black. In the right figure, $-\log_{10}(p)$ values are presented for the \texttt{Freesurfer-WALD} (left), \texttt{Freesurfer-RCoT} (center), and \texttt{Fastsurfer-RCoT} (right) across traits with all brain structures. Unicolored points and triangles represent curiosity, diligence, nervousness, neuroticism, sociability and warmth (left to right), while grey-filled points and triangles represent DNCITs for all BFBTs together. In both figures, tests with $-\log_{10}(p)>2.5$ are colored red and annotated. The solid vertical lines depict the $-\log_{10}(\alpha_{\text{bonf}})$ where $\alpha_{\text{bonf}}$ is the significance level Bonferroni adjusted  for 642 (left) and 6 (right) tests for an individual significance level of 0.05, i.e. $\alpha_{\text{bonf}}=\frac{0.05}{642}$ and $\alpha_{\text{bonf}}=\frac{0.05}{6}$, respectively.}
\label{fig:p-values}
\end{figure}

We depict the results for individual brain structures and BFBTs on the left in Figure \ref{fig:p-values}. The traits are shown on the horizontal axis and for each trait, 107 p-values are obtained, one for each brain measure. The five lowest p-values in \citet[sec.~3.2]{avinun2020little} are colored in black. Additionally, all p-values with $-\log_{10}(p)>2.5$ are colored in red and annotated with their corresponding brain measure. 

On the right in Figure \ref{fig:p-values}, the results of the \texttt{Freesurfer-WALD} applied to all brain measures jointly are shown for each BFBT in the left column. The order of the BFBTs corresponds to the order of the BFBTs on the left in Figure \ref{fig:p-values}. In the center, the p-values of the \texttt{Freesurfer-RCoT} applied to all brain measures for each trait as well as all BFBTs jointly as larger point are shown. Finally, the right column depicts the corresponding p-values for the \texttt{Freesurfer-RCoT} and each BFBT, as well as all BFBTs jointly, as triangles. 

We highlight three main findings from the analysis. First, after multiple testing adjustment with the Bonferroni method, all p-values of individual tests are above the threshold $\alpha_{\text{bonf}}$, agreeing with the little evidence found for associations between brain morphometry measures and BFBT proxies in healthy individuals in \citet{avinun2020little}, but now confirmed in the larger cohort of the UKB. 
The only significant p-value after multiple testing adjustment at a significance level of 0.1 is for the posterior thalamic radiation and neuroticism.
Second, we observe that the p-values in the more powerful tests \texttt{Freesurfer-RCoT} and \texttt{Fastsurfer-RCoT} are larger compared to the Wald-based tests. Our simulation study, which indicated inflated T1E for Wald-based tests, suggests that the smaller p-values of the Wald-tests may only be due to not sufficiently controlling for the nonlinear confounding effects with the confounder set used here.
Third, the \texttt{Freesurfer-RCoT} and \texttt{Fastsurfer-RCoT} for the brain imaging data and all behavioral traits are not significant. We expect this test to adequately account for all nonlinear relationships with confounding variables, as shown in the simulation study. 
Thus, we conclude that there is no evidence for links between the brain structure as measured by brain MRI scans and personality traits in healthy individuals, also in the larger UKB cohort and using our new more powerful CITs.

\subsection{Confounder Control}\label{subsubsec:confounder}

The goal of confounder control in hypothesis testing is to control for their effects so that significant associations are not due to associations by the confounders.
Thus, the way confounder control is conducted significantly affects the validity of tested associations between psychological traits and brain measures, with inadequate control potentially leading to spurious correlations \citep{hyatt2020quandary, alfaro2021confound}. 
A common method of confounder control is to linearly regress $X$ and $Y$ on the confounders $Z$ and potentially some pre-defined nonlinear transformations and interaction terms of $Z$ ('regress out'). 
The resulting residuals are assumed to be controlled for the confounding effects and are used in independence tests.
However, there may still be uncontrolled nonlinear and interaction effects that lead to spurious correlations.
DNCITs can be used to test whether regressing out does sufficiently control for nonlinear and interaction effects of individual confounders or entire confounder sets in imaging data using algorithm \ref{alg:conf_control}. 
We illustrate its use on the conventional confounder set for brain imaging data used throughout the previous subsection.
The presence of insufficiently controlled effects in the imaging data may result in spurious correlations if the confounder remains additionally conditionally associated with $Y$ after regressing it out.
\begin{algorithm}
\caption{Test for sufficient control of the effects of confounders on imaging data via regress out}\label{alg:conf_control}
\KwIn{Confounder set $\mathcal{C}$, feature representations $X^\omega$}
\KwOut{p-values for uncontrolled confounding effects of all confounders $Z_i\in \mathcal{C}_{base}$}
\BlankLine
Initialize an empty list \textit{p-values} to store the p-values\;
\For{$Z_i\in \mathcal{C}_{base}$}{
\begin{enumerate}
    \item Linearly regress $X^\omega$ on the confounder $Z_i$ and possibly its interactions and \\
    nonlinear transformations $l(Z_i)$ from $\mathcal{C}$; 
    obtain 
    the corresponding residuals \\
    $X^{\omega, res}$
    \item Test with a nonparametric CIT $\varphi_\theta^\omega$
    \begin{align*}
H_0^\omega: X^{\omega, \text{res}} \indep Z_i \mid \mathcal{C} \setminus \{Z_i,l(Z_i)\} \quad \text{vs} \quad H_1^\omega: X^{\omega, \text{res}} \not\indep Z_i \mid \mathcal{C} \setminus \{Z_i,l(Z_i)\}.
\end{align*}
 \item Store the p-value from the test in \textit{p-values}.
\end{enumerate}
}
\Return{List p-values.}
\end{algorithm}

To test for sufficient control of the effects of each of the confounders in the conventional confounder set on the brain MRIs, we use the \texttt{Residual-RCoT}. This method is similar to the \texttt{Deep-RCoT} but involves regressing out a confounder, its pre-specified interaction terms, and nonlinear transformations from the feature representation before applying the RCoT. This allows us to test whether effects from a particular confounder can still be detected, given the remaining conventional confounder set. In particular, we denote the confounder set consisting of the confounders, their nonlinear and interaction terms as $\mathcal{C}$, and the set consisting of the confounders only as $\mathcal{C}_{base}$. We first regress out one confounder and potentially its nonlinear and interaction terms as defined in the conventional confounder set $\mathcal{C}$ in subsection \ref{subsec:brain_behaviour_data_descrip} from the brain measures. Then, we use the RCoT to test for conditional associations between the residuals and the confounder from $\mathcal{C}_{base}$, given all other confounders and their nonlinear and interaction terms in the conventional confounder set $\mathcal{C}$. Applying this algorithm to each confounder in $\mathcal{C}_{base}$, we obtain the following p-values: age $6.7\times 10^{-2}$; sex $3.16\times10^{-16}$; 5 genetic PCs $2.5\times 10^{-1}$; head size $2.59\times10^{-6}$; head position $1.9\times 10^{-2}$; assessment date $8.4\times 10^{-3}$; assessment site $2.3\times 10^{-1}$. These results indicate particularly strong effects of sex and head size, even after regressing out these confounders from the brain MRIs and controlling for all other variables in the conventional confounder set. This insufficiently controlled effects in the imaging data may result in spurious correlations if the confounder remains also conditionally associated with the behavioral traits after regressing it out.

This finding is consistent with the results of \citet{alfaro2021confound}, who showed that regressing out the conventional confounder set controls for only part of the variation in brain measures due to confounders. While \citet{alfaro2021confound} explored the approach of expanding the confounder set by including additional nonlinear and interaction terms, we highlight the possibility of testing for more complex relationships than allowed for in parametric regress out or Wald tests using our DNCITs. Our results in the simulation study and the application in section \ref{sec:sim} and subsection \ref{subsec:brain_behavior} then indicate that the \texttt{Deep-RCoT} controls for confounding effects more adequately than regressing out the conventional confounder set.

\section{Conclusion}\label{sec:conclusion}

We introduced DNCITs to test for conditional associations between an image and a scalar given vector-valued confounders, providing a theoretically sound framework for CI testing in this context.
We presented theoretical results showing the validity of DNCITs using embedding maps, and provided sufficient conditions on the learned embedding map for these to hold, which are fulfilled in addition to sample splitting in particular by transfer and (conditional) unsupervised learning. The latter approaches eliminate the need for sample splitting, leading to potentially more powerful tests and significantly reducing computational costs by utilizing already available embedding maps.
Furthermore, we investigated the performance of the DNCITs both theoretically and empirically in a novel simulation design in dependence of the chosen embedding maps and nonparametric CITs. 
Moreover, we demonstrated the usefulness of DNCITs in two real-world applications, first in extending recent studies in personality neuroscience and second in testing for sufficient confounder control in brain imaging data. 
Our empirical results identified nonparametric CITs with both Type I error control and relatively high power, depending on the sample size.
In addition, they highlighted the critical role of confounder control, revealing also the limitations of current practices that rely on parametric tests and their confounder control. 
Finally, we provide an R-package that provides a choice between several nonparametric CITs as well as useful default embedding maps in particular being provided for 2D images.

While we here focused on a modular framework that mainly works with existing embedding maps, which has advantages in particular for small sample sizes, future work could look at tests that optimally learn the embedding map to maximize power, but would need to tackle challenges of post-selection inference.

In conclusion, DNCITs enable statistical testing, a key inference tool, for commonly analyzed multimodal data sets of images and tabular data, such as in the biomedical domain. While our focus has been on images, the theoretical framework can also be applied to other multimodal data types, including text.

\acks{Marco Simnacher, Xiangnan Xu, Hani Park, Christoph Lippert and Sonja Greven were 
funded by the Deutsche Forschungsgemeinschaft (DFG, German Research Foundation) - project number 459422098. 
This research has been conducted using the UK Biobank Resource under Application Number 77717.
}

\bibliography{references}

\appendix
\clearpage
\newpage

\section{Proofs}\label{app:proofs}
\printProofs

\section{Conditional Independence after Transformations}
\label{app:CIT_after_transformations}

This section discusses a generalization of Theorem \ref{thm:null_hypotheses} which allows for transformations of $X$ and $Y$.  Therefore, let $(\mathfrak{Y},\mathcal{F}_\mathfrak{Y}), (\mathcal{E}, \mathcal{F}_\mathcal{E})$ be measurable spaces, $\hat{\gamma}$ be a $(\mathcal{E},\mathcal{F}_\mathcal{E})$-valued random variable and $\rho:\mathcal{Y}\times \mathcal{E}\to \mathfrak{Y}$ be a measurable function of $Y$ and $\hat{\gamma}$, and denote $Y^\rho=\rho(Y,\hat{\gamma})$. Based on this, we introduce the hypotheses
\begin{align}\label{test_hypotheses_two_transformations}
    H_0^{\omega,\rho}: X^\omega\indep Y^\rho|Z\quad vs.\quad H_1^{{\omega,\rho}}: X^\omega\notindep Y^\rho|Z.
\end{align}
Then, $X\indep Y|Z$ implies $X^\omega\indep Y^\rho|Z$ under analogous assumptions on the parameters $\hat{\beta}$ and $\hat{\gamma}$ of the embedding maps $\omega$ and $\rho$ as in Theorem \ref{thm:null_hypotheses}:
\begin{theoremE}
[Relation of null hypotheses $H_0$ and $H_0^{\omega,\rho}$ after transformations][normal]\label{thm:null_hypotheses_transformations} Let $(\mathfrak{X}, \mathcal{F}_\mathfrak{X}), (\mathfrak{Y},\mathcal{F}_\mathfrak{Y}),(\mathcal{B},\mathcal{F}_\mathcal{B}), (\mathcal{E},\mathcal{F}_\mathcal{E})$ be measurable spaces. Furthermore, let $\hat{\beta}$ be a $(\mathcal{B},\mathcal{F}_\mathcal{B})$-valued random variable such that $\hat{\beta}$ is conditionally independent of $Y$ given $X,Z$, and  let $\omega:\mathcal{X}\times \mathcal{B}\to \mathfrak{X}$ be a measurable function of $X$ and $\hat{\beta}$ and denote $X^{\omega}=\omega(X,\hat{\beta})$. Moreover, let $\hat{\gamma}$ be a $(\mathcal{E},\mathcal{F}_\mathcal{E})$-valued random variable such that $\hat{\gamma}$ is conditionally independent of $X^\omega$ given $Y,Z$, and  let $\rho:\mathcal{Y}\times \mathcal{E}\to \mathfrak{Y}$ be a measurable function of $Y$ and $\hat{\gamma}$ and denote $Y^{\rho}=\rho(Y,\hat{\gamma})$. \\
If $H_0:X\indep Y|Z$ is true, then $H_0^{\omega, \rho}:X^{\omega}\indep Y^\rho|Z$ is true.    
\end{theoremE}
\begin{proofE}
    By Theorem \ref{thm:null_hypotheses} it follows that $X^\omega\indep Y|Z$. Furthermore, by \cite[Lemma~4.3]{dawid1979conditional}, for $X^\omega\indep Y|Z$, $X^\omega\indep \hat{\gamma} |(Y,Z)$, it follows that
    \begin{align*}
        X^\omega\indep (Y,\hat{\gamma})|Z.
    \end{align*}
    Since $\rho$ is a measurable function of $(Y,\hat{\gamma})$, we have by \cite[Lemma~4.2]{dawid1979conditional} that $X^\omega\indep \rho(Y,\hat{\gamma})|Z$, i.e. $X^{\omega}\indep Y^\rho|Z$.
\end{proofE}
This is more general than formulations such as in \citet{dawid1979conditional}, since we not only allow for functions $\widetilde{\omega}:\mathcal{X}\to \mathfrak{X}$ of $X$, but instead quantify exactly the additional information that the parameters $\hat{\beta}$ of the embedding maps $\omega:\mathcal{X}\times\mathcal{B}\to \mathfrak{X}$ are allowed to use (analogously for $\rho$) to lead to conditionally independent feature representations under the null hypothesis \eqref{test_hypotheses_image}. In particular, $\hat{\beta}$ can be obtained from all information except that in $Y$, given $X, Z$. Thus, we allow the embedding maps to use information in $Z$ and $X$ or $Y$, respectively, for $\omega$ and $\rho$, respectively. We discussed learning methods for embedding maps that use only such information in and after Corollary \ref{corollary:embeddings}.

\section{Simulation Study Details}\label{app:sim_study}

\begin{figure}
    \centering    
    \includegraphics[width=0.8\textwidth]{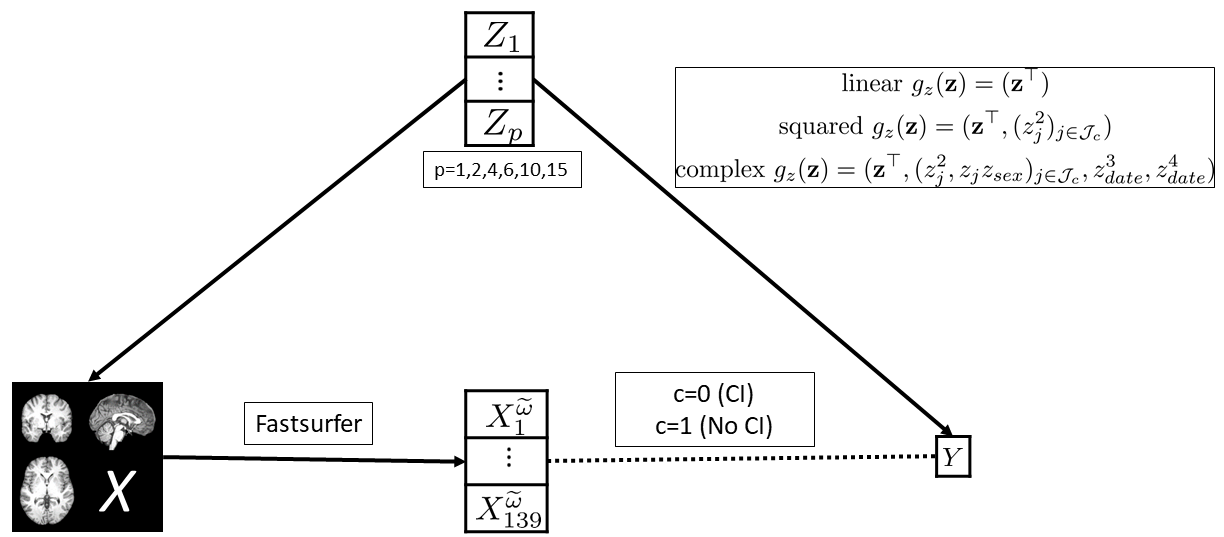}
    \caption{Illustration of the 18 data-generating mechanisms (DGMs) used in the simulation study. Solid black lines represent the relationships between variables, while the dotted black line indicates either conditional independence ($c=0$) or dependence ($c=1$). The feature representation, denoted by $X^{\widetilde{\omega}} = (X^{\widetilde{\omega}}_1, \dots, X^{\widetilde{\omega}}_{139})$, is derived from the Fastsurfer embedding map and is used to simulate the conditional (in)dependence between $X$ and $Y$. The confounders $Z$ are connected to the brain MRI scans based on their real-world associations in the UK Biobank (UKB). We vary the confounder dimension $p$ across 1, 2, 4, 6, 10, and 15 while fixing the confounder relationship to $g_z(\mathbf{s})=(\mathbf{s}^\top,(s_j^2)_{j\in\mathcal{J}_c})$. Additionally, the confounder relationship is varied over $g_z(\mathbf{s})= \mathbf{s}^\top, g_z(\mathbf{s})=(\mathbf{s}^\top,(s_j^2)_{j\in\mathcal{J}_c}),$ and $g_z(\mathbf{s})=(\mathbf{s}^\top,(s_j^2, s_{j}s_{sex})_{j\in\mathcal{J}_c},s_{date}^3,s_{date}^4)$ while the confounder dimension is fixed to $6$. $Y$ is generated according to Equation \eqref{equ:sim_setting}, as a linear combination of $X^{\widetilde{\omega}}$ and $Z$ transformed by one of the functions $g_z$ ($\mathcal{J}_c$ denotes the index set of continuous variables).}
    \label{fig:dgms}
\end{figure}

\begin{figure}
\centering    
    \includegraphics[width=0.65\textwidth]{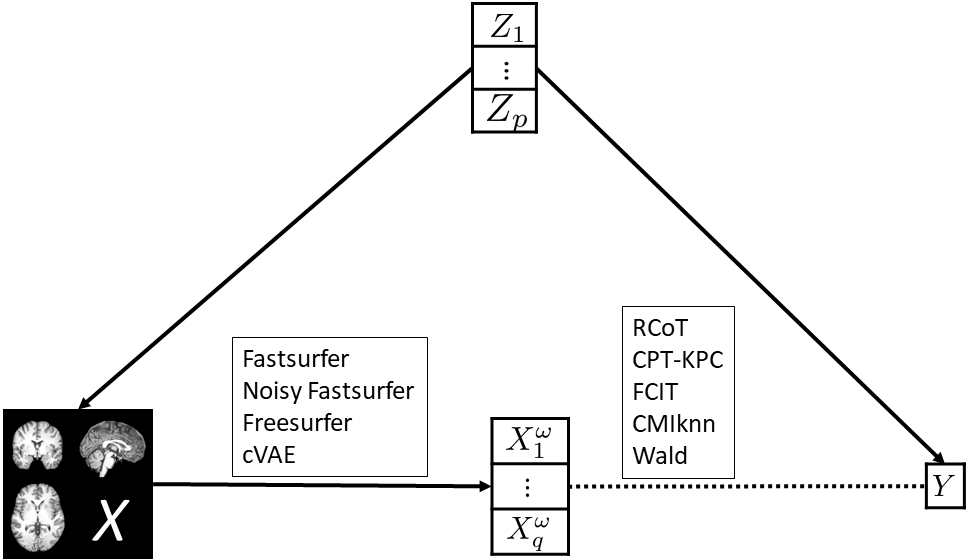}
    \caption{Illustration of the 20 DNCITs applied in the simulation study. Solid black lines represent the relationships between variables, while the dotted black line indicates conditional (in)dependence to be tested. For each (nonparametric) CIT, we apply the true Fastsurfer embedding map $(X_1^\omega,\hdots,X^{\omega}_{139})$, a noisy version of it $(X_1^\omega,\hdots,X_{139}^\omega)$, the Fastsurfer embedding map $(X_1^\omega,\hdots,X_{165}^\omega)$, and a conditional VAE embedding map $(X_1^\omega,\hdots,X_{256}^\omega)$. The (nonparametric) CITs are varied over the RCoT, CPT-KPC, CMIknn, FCIT and the Wald test.}
    \label{fig:dncits_sim}
\end{figure}

An additional illustrative overview over all simulation designs and DNCITs can be found in the Figures \ref{fig:dgms} and \ref{fig:dncits_sim}.

\subsection{Simulations' Parameter Choices}\label{app:sim_realism}

The sample sizes are chosen to be approximately linear on the logarithmic scale, as in \citet{marek2022reproducible}, where the authors used these sample sizes 
in a study on the reproducibility of significant findings in brain-wide association studies (BWAS). We also include $n=5000, 10000$ to account for the large number of observations available in the UKB.

\textbf{Regarding the DGMs:} We repeatedly resample pairs $(X,Z)$ from the UKB to obtain samples with the true association of $Z$ and $X$ in the UKB. This makes the results both realistic and particularly relevant to the UKB brain imaging data, but one should be careful about extrapolating the results to other dataset, as the associations between $X$ and $Z$ may differ. In addition, the choice to simulate the conditional dependence between $X$ and $Y$ by the Fastsurfer feature representations $\mathbf{X}^{\widetilde{\omega}}$ makes the results particularly relevant to brain morphometry measures obtained e.g. by the Fastsurfer or the Freesurfer pipeline \citep{fischl2012freesurfer, alfaro2018image, henschel2020fastsurfer}, which are often used in BWAS \citep{hyatt2020quandary, marek2022reproducible}.

\textbf{Regarding the embedding maps of the DNCITs:} The true Fastsurfer feature representations, used to model the conditional dependence, consist of volume measurements of grey matter in different parts of the brain. The noisy Fastsurfer feature representation mimics a noisy recovery of these true feature representations. The Freesurfer embedding map is closely related to the true feature representation, as it also measures brain structures through volume measurements in regions included in the true representation. However, not all volume measurements in the Fastsurfer feature representation are present in the Freesurfer representation, which additionally includes additional area, volume, and thickness measurements. Therefore, the Freesurfer embedding map represents a feature set with slightly different and additional information. The cVAE, trained on the ADNI data set, and its feature representations can be interpreted as an unsupervised transfer learning method to obtain feature representations for the DNCITs. The four feature representations represent decreasing similarity to the true Fastsurfer feature representation.

In summary, the simulation study represents settings following findings from large studies on brain imaging data \citep{hyatt2020quandary, alfaro2021confound, marek2022reproducible}, especially in terms of the confounding effects.
Finally, the code and the introduced designs can be used to evaluate the performance of the DNCITs on other imaging datasets before applying them to the corresponding real-world application.

\subsection{Detailed Results}\label{app:detailed_sim}
 \begin{measuredfigure}
\centering
    \includegraphics[width=0.85\linewidth]{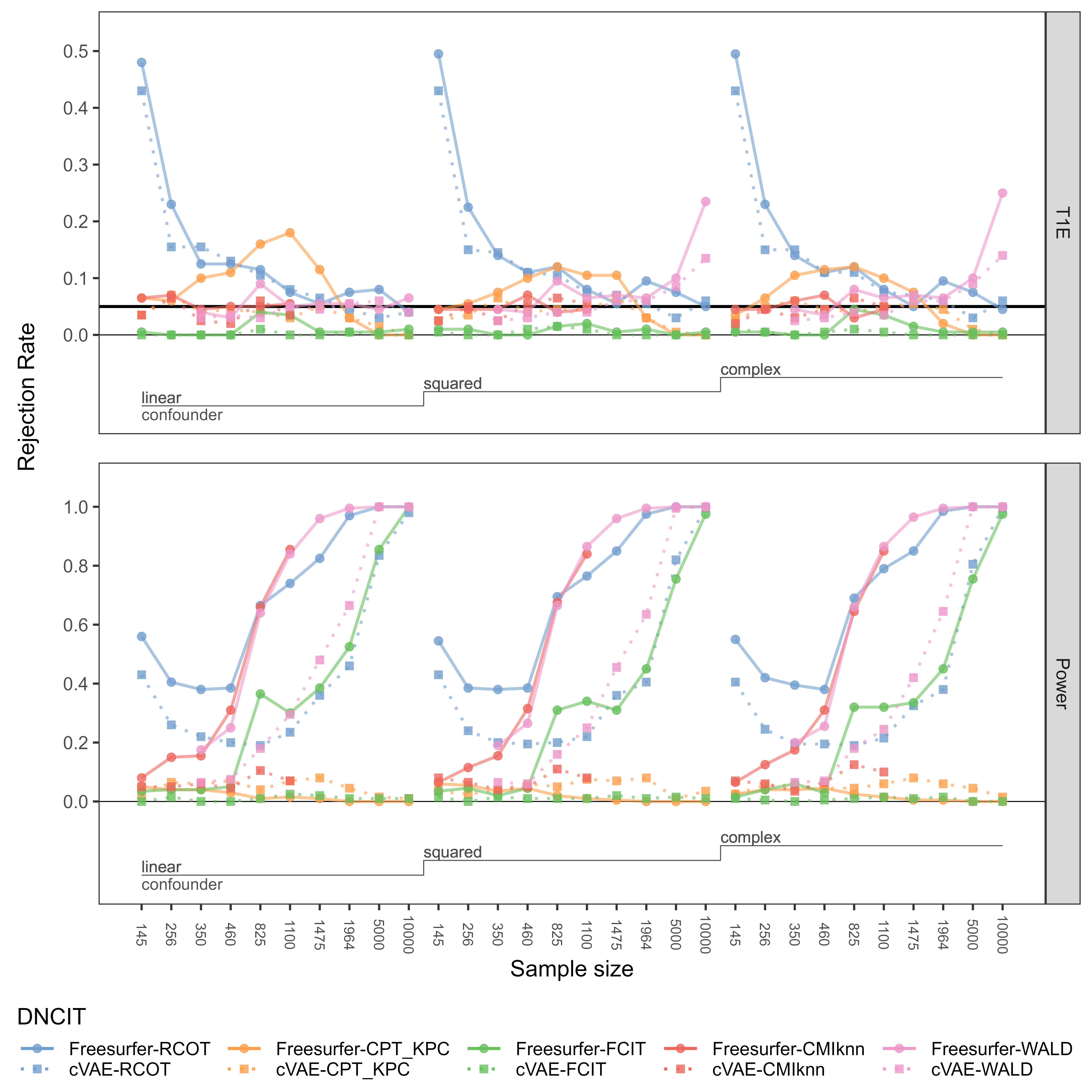}
    \caption{The rejection rates of the DNCITs for $c=0$ (CI, top) and $c=1$ (no CI, bottom) for increasingly complex confounder relations (columns). For each column, the sample size is increased from left to right. The confounder dimension is set to 6. Horizontal lines at 0 and $\alpha=0.05$.}
    \label{fig:sim_conf_relation_detailed}
    \end{measuredfigure}

\begin{landscape}
\begin{figure}
    \centering
    \includegraphics[width=\linewidth]{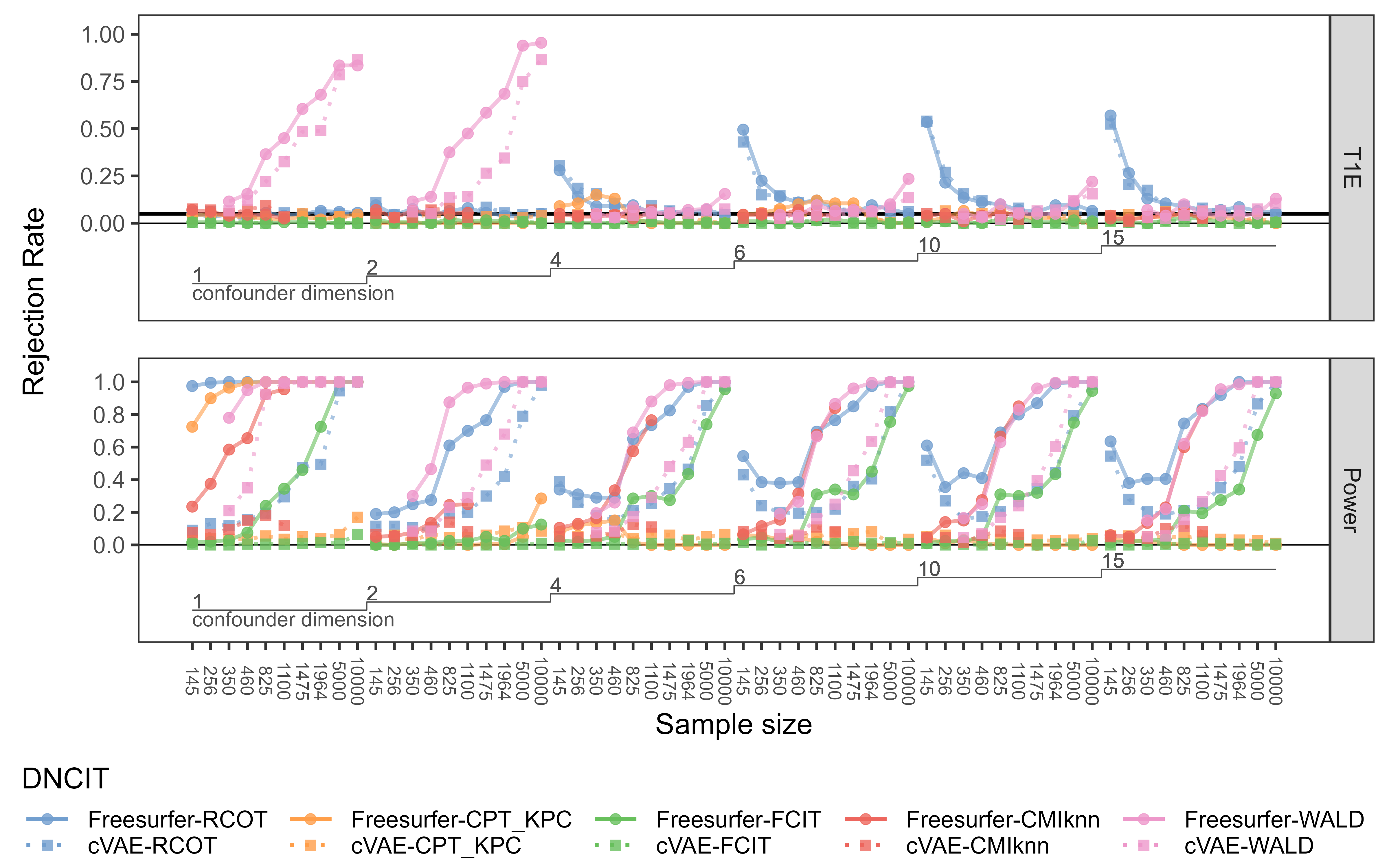}
    \caption{The rejection rates of the DNCITs for $c=0$ (CI, top) and $c=1$ (no CI, bottom) for increasing confounder dimension (columns). For each column, the sample size is increased from left to right. The confounder relationship is set to $g_z(\mathbf{s})=(\mathbf{s}^\top,(s_j^2)_{j\in\mathcal{J}_c})$, where $\mathcal{J}_c$ denotes the index set of continuous variables. Horizontal lines at 0 and $\alpha=0.05$.}
    \label{fig:sim_conf_dim_detailed}
\end{figure}
\end{landscape}

\begin{figure}
    \centering
    \includegraphics[width=\linewidth]{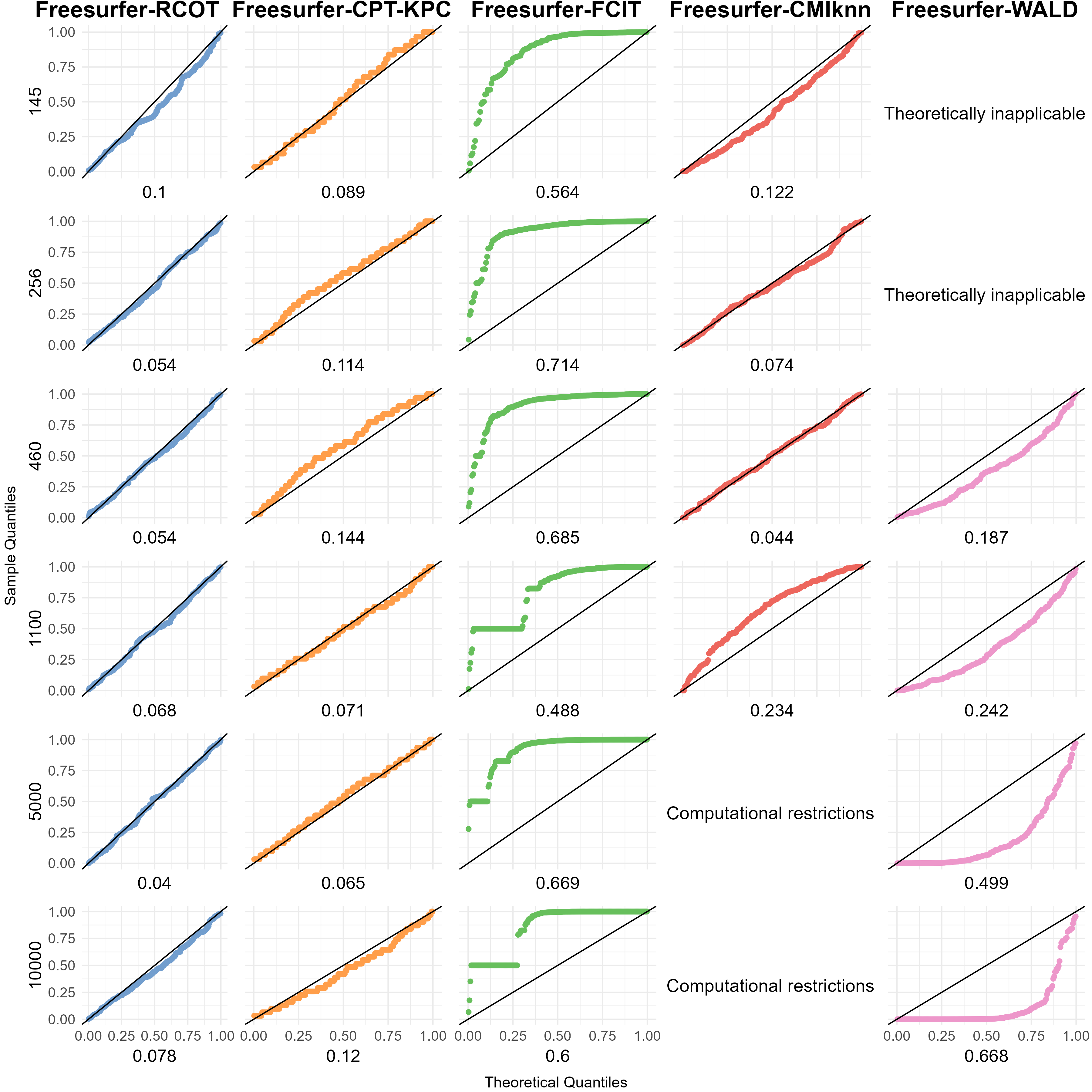}
    \caption{QQ-plots of the observed p-values against the theoretical quantiles of a uniform distribution on the interval $[0,1]$ for all nonparametric CITs applied to the Freesurfer embedding map at selected sample sizes (indicated at the y-axis label). The results correspond to the DGM under conditional independence with one confounder and the quadratic confounder relationship $g_z(\mathbf{s})=(\mathbf{s}^\top,(s_j^2)_{j\in\mathcal{J}_c})$, where $\mathcal{J}_c$ denotes the index set of continuous variables. The diagonal black lines $y=x$ serve as references for the theoretical quantiles of the uniform distribution on $[0,1]$. If the p-values are uniformly distributed, they should align along this line. The x-axis labels display the Kolmogorov-Smirnov test statistics used to assess whether the p-values are uniformly distributed; smaller values indicate a better calibration of the CIT, as the DGM assumes conditionally independence.}
    \label{fig:qq-plot_dim1_squared_freesurfer_ci}
\end{figure}

\begin{figure}
    \centering
    \includegraphics[width=\linewidth]{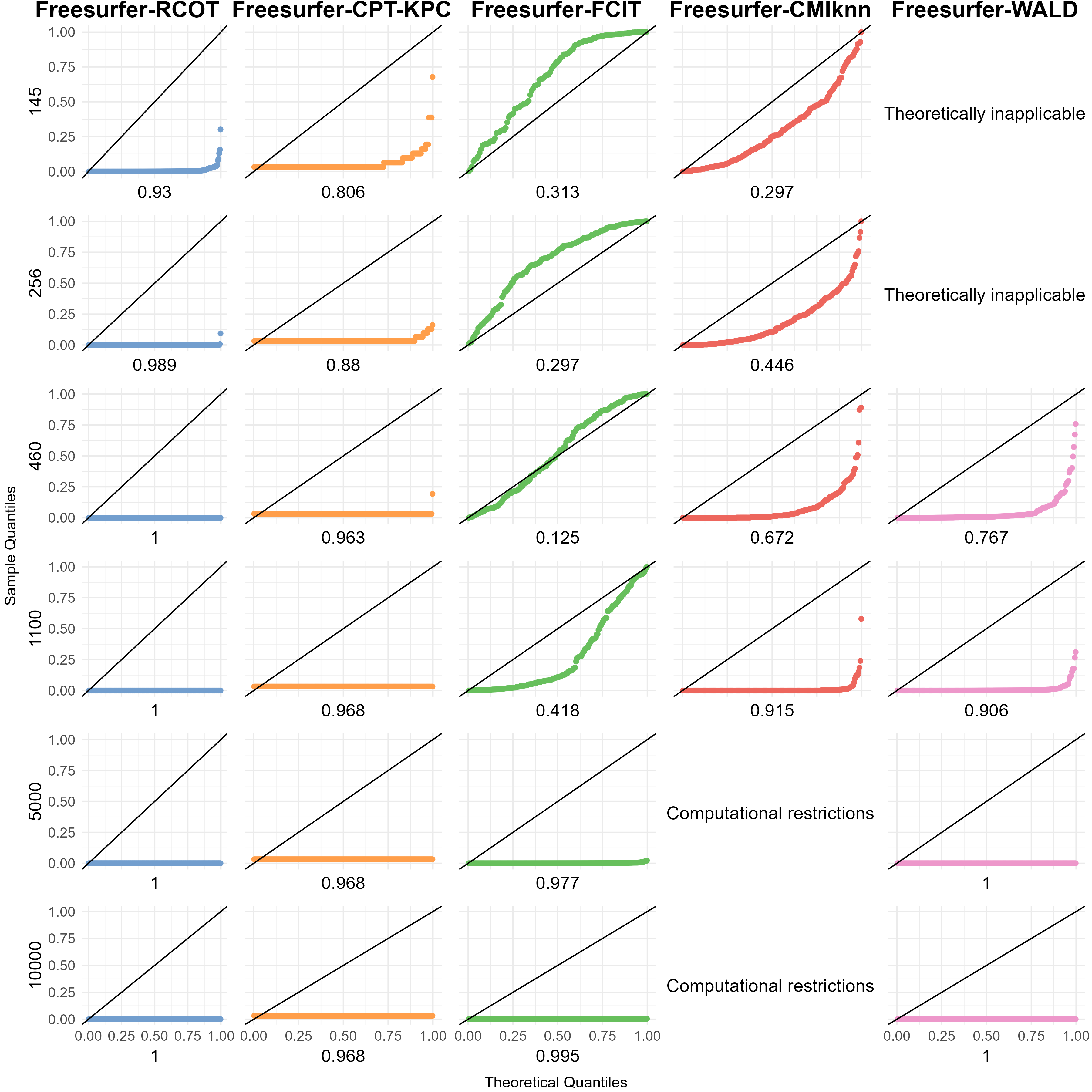}
    \caption{QQ-plots of the observed p-values against the theoretical quantiles of a uniform distribution on the interval $[0,1]$ for all nonparametric CITs applied to the Freesurfer embedding map at selected sample sizes (indicated at the y-axis label). The results correspond to the DGM under conditional dependence with one confounder and the quadratic confounder relationship $g_z(\mathbf{s})=(\mathbf{s}^\top,(s_j^2)_{j\in\mathcal{J}_c})$, where $\mathcal{J}_c$ denotes the index set of continuous variables. The diagonal black lines $y=x$ serve as references for the theoretical quantiles of the uniform distribution on $[0,1]$. If the p-values are uniformly distributed, they should align along this line. The x-axis labels display the Kolmogorov-Smirnov test statistics used to assess whether the p-values are uniformly distributed; values close to zero indicate a bad calibration of the CIT, as the DGM assumes conditionally dependence.}
    \label{fig:qq-plot_dim1_squared_freesurfer_no_ci}
\end{figure}

\begin{figure}
    \centering
    \includegraphics[width=\linewidth]{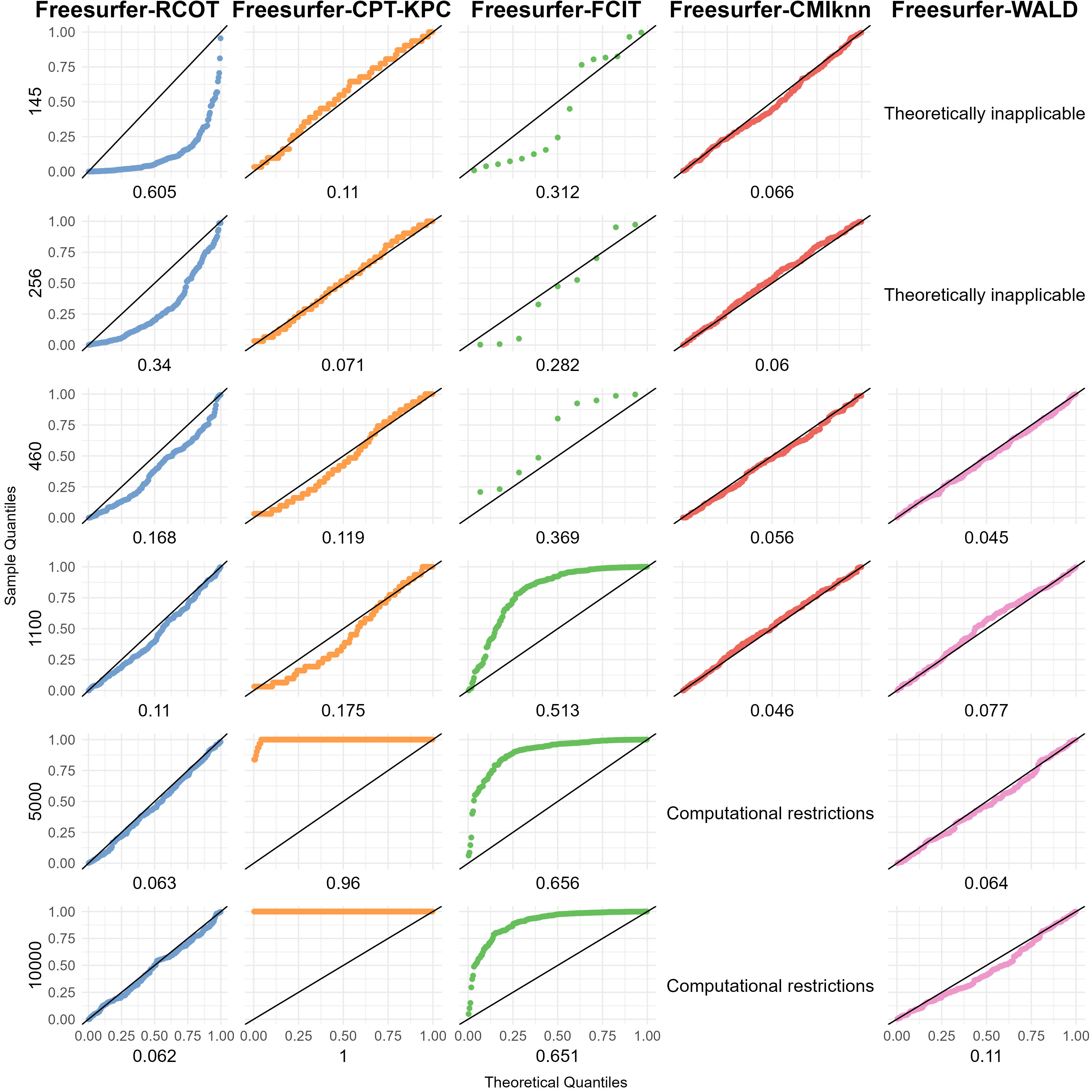}
    \caption{QQ-plots of the observed p-values against the theoretical quantiles of a uniform distribution on the interval $[0,1]$ for all nonparametric CITs applied to the Freesurfer embedding map at selected sample sizes (indicated at the y-axis label). The results correspond to the DGM under conditional independence with six confounders and the quadratic confounder relationship $g_z(\mathbf{s})=(\mathbf{s}^\top,(s_j^2)_{j\in\mathcal{J}_c})$, where $\mathcal{J}_c$ denotes the index set of continuous variables. The diagonal black lines $y=x$ serve as references for the theoretical quantiles of the uniform distribution on $[0,1]$. If the p-values are uniformly distributed, they should align along this line. The x-axis labels display the Kolmogorov-Smirnov test statistics used to assess whether the p-values are uniformly distributed; smaller values indicate a better calibration of the CIT, as the DGM assumes conditionally independence. The \texttt{Freesurfer-FCIT} can produce errors for sample sizes up to 460 due to the implementation of its prediction models, resulting in fewer p-values in the corresponding grids.}
    \label{fig:qq-plot_dim6_squared_freesurfer}
\end{figure}

\begin{figure}
    \centering
    \includegraphics[width=\linewidth]{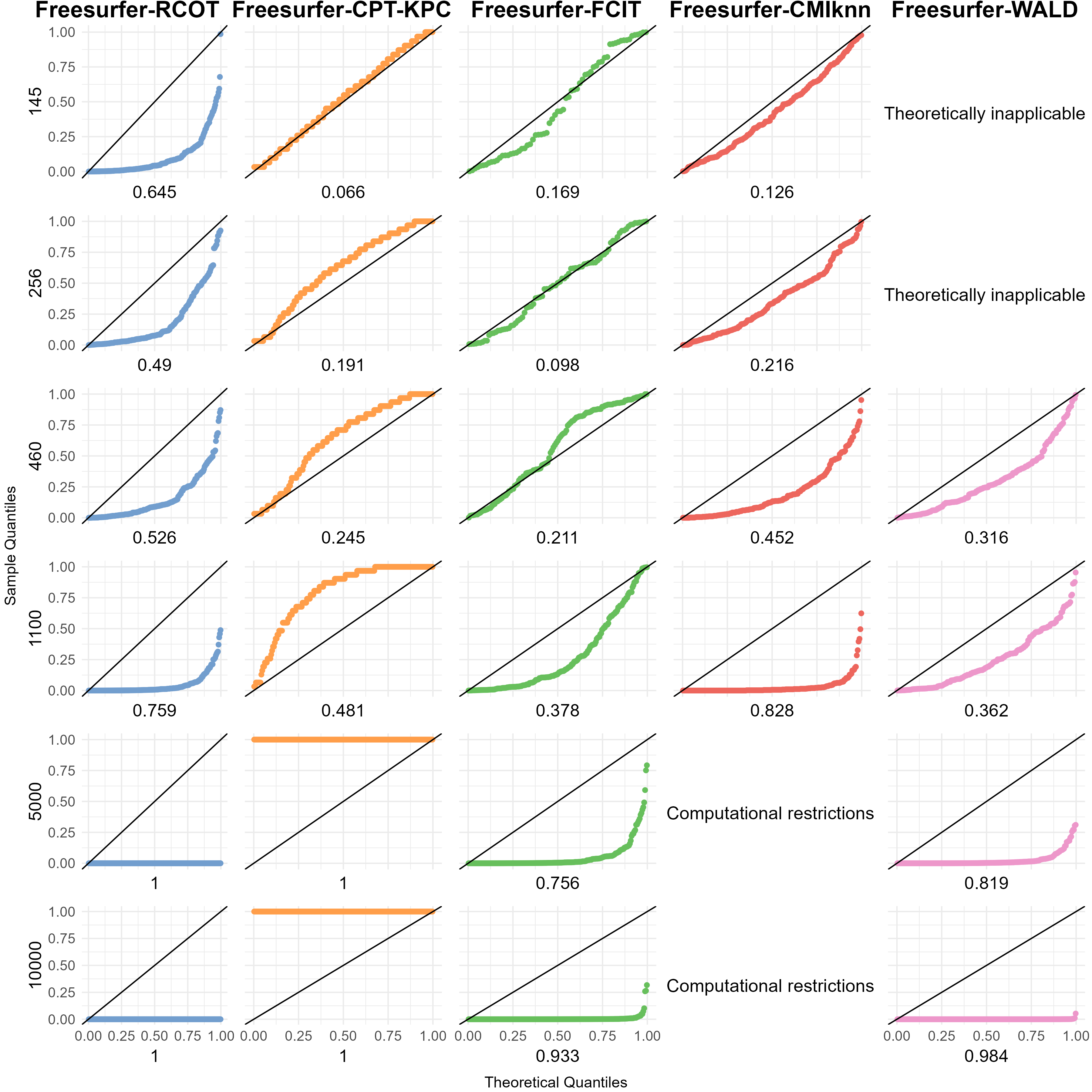}
    \caption{QQ-plots of the observed p-values against the theoretical quantiles of a uniform distribution on the interval $[0,1]$ for all nonparametric CITs applied to the Freesurfer embedding map at selected sample sizes (indicated at the y-axis label). The results correspond to the DGM under conditional dependence with six confounders and the quadratic confounder relationship $g_z(\mathbf{s})=(\mathbf{s}^\top,(s_j^2)_{j\in\mathcal{J}_c})$, where $\mathcal{J}_c$ denotes the index set of continuous variables. The diagonal black lines $y=x$ serve as references for the theoretical quantiles of the uniform distribution on $[0,1]$. If the p-values are uniformly distributed, they should align along this line. The x-axis labels display the Kolmogorov-Smirnov test statistics used to assess whether the p-values are uniformly distributed; values close to zero indicate a bad calibration of the CIT, as the DGM assumes conditionally dependence.}
    \label{fig:qq-plot_dim6_squared_freesurfer_no_ci}
\end{figure}

\beginsupplement
\section{Details, Adaption and Implementation of Selected Conditional Independence Tests}\label{app:online}

First, we give a short summary of the selected nonparametric CITs and analyse them afterwards in more detail w.r.t. our setting. 
We construct a nonparametric CIT using the KPC from \citet{huang2022kernel} as the test statistic together with the CPT from \citet{berrett2020conditional, spisak2022statistical}. While the kernel of the KPC adapts well to the dimension of the feature representations and its geometric graph functionals to the dimension of the confounders, the CPT framework ensures the validity of the test. In addition, the KPC allows a flexible choice of kernel and is implemented in a stable manner. Finally, ancillary data for $\mathbb{P}^{Y|Z}$ is often available, especially in the context of medical imaging, which can be expensive for many patients, while tabular measurements may be more readily available in larger quantities, such as in the UKB \citep{sudlow2015uk}.
In addition, we select the RCoT of \citet{strobl2019approximate} because of its fast approximation of the KCIT with similar performance, its stable implementation, and the often strong performance of kernel-based methods in high-dimensional settings.
Furthermore, we choose the prediction-based CITs FCIT of \citet{chalupka2018fast}
because of its adaptation to the different dimension between the feature representations of the image $X$ and the scalar $Y$.
Finally, the CMIknn of \citet{runge2018conditional} is chosen to represent metric-based CITs. 

\subsection{The Conditional Permutation Test and the Kernel Partial
Coefficient}\label{app:cpt-kpc}

Throughout our empirical results, we approximate for CPT-based CITs $\mathbb{P}^{Y|Z_i=\mathbf{z}_i}$ by a normal distribution with mean and variance estimated by a generalized additive model in sample,
where the effect of each continuous confounder in $Z_i=(Z_{i1},\hdots,Z_{ip})$ is modeled by a smooth term and that of each categorical confounder with a separate parameter for each category.\footnote{The normal distribution could be replaced by other distribution families available for generalized additive models. Similar flexibility holds for the modelling of the continuous and categorical confounder.} This is analogous to the approach in \citet{spisak2022statistical}, however we use in the implementation the more reliable R-package mgcv from \citet{wood2015package} compared to the python-package pyGAM from \citet{danielserven20181208723} used in \citet{spisak2022statistical}. Nevertheless, the method to approximate $\mathbb{P}^{Y|Z=\mathbf{z}_i}$ is adaptable and depends on $\mathcal{Z}$ and $\mathcal{Y}$ as well as on potential supplementary data $(Y_i,Z_i)_{i=n+1,\hdots,n+n'}$.

The time complexity of CPT-based CITs is the time complexity of approximating $\mathbb{P}^{Y|Z}$ plus $M$ times the time complexity of estimating the test statistic, where $M$ is the number of permutations of $Y$. For example, a detailed description of the time complexity for generalized additive models and their extensions can be found in \citet{wood2020inference}, and in our specific implementation is given by $\mathcal{O}(ns^2)$, where $s$ is the number of parameters in the basis representation of $Z$ used to approximate $\mathbb{P}^{Y|Z}$ and $n$ is the sample size of the data set. However, this could be reduced using different implementations, and we observe throughout the simulation study that for the KPC test statistic that we consider, estimating the test statistic is the bottleneck in terms of computational cost.

Together with the CPT, we have chosen the conditional dependence measure recently presented in \citet{huang2022kernel} as test statistic, which extends upon measures established in \citet{deb2020measuring, azadkia2021simple}. Specifically, they propose the kernel partial correlation (KPC) as 
\begin{align}\label{equ:test_stat_KPC}
    KPC_\theta((X^\omega,Y,Z))\coloneqq \frac{\mathbb{E}[MMD^2(\mathbb{P}^{X^\omega|YZ},\mathbb{P}^{X^\omega|Z})]}{\mathbb{E}[MMD^2(\delta_{X^\omega},\mathbb{P}^{X^\omega|Z})]}
\end{align}
where $MMD$ denotes the maximum mean discrepancy as for example defined in \citet{gretton2012kernel}, $\delta_{X^\omega}$ is the Dirac measure at $X^\omega$ and $\theta$ are the parameters of the kernels introduced below.

For a characteristic, finite kernel $k_{\mathfrak{X}}$, separable RKHS $\mathcal{H}_\mathfrak{X}$, and the assumption that $X^\omega$ is not a measureable function of $Z$, it holds that $KPC_\theta((X^\omega,Y,Z))\in[0,1]$ and in particular, $KPC_\theta((X^\omega,Y,Z))=0$ if and only if $X^\omega\indep Y|Z$. Thus, together with the CPT, the hypothesis \eqref{test_hypotheses_embedding} can be tested, under the above assumptions equivalently, by testing
\begin{align*}
    H_{0,KPC}^\omega:KPC_\theta((X^\omega,Y,Z))=0\quad vs. \quad H_{1,KPC}^\omega:KPC_\theta((X^\omega,Y,Z))>0. 
\end{align*} 
Furthermore, the authors showed in \citet[Lemma~1,2]{huang2022kernel} that the KPC is well defined and equivalent to
\begin{align*}
    KPC_\theta((X^\omega,Y,Z))=\frac{\mathbb{E}[\mathbb{E}[k_{\mathfrak{X}}(X^\omega_2,X^{\omega'}_2)|Y,Z]]-\mathbb{E}[\mathbb{E}[k_{\mathfrak{X}}(X^\omega_1,X^{\omega'}_1)|Z]]}{\mathbb{E}[k_{\mathfrak{X}}(X^\omega, X^\omega)]- \mathbb{E}[\mathbb{E}[k_{\mathfrak{X}}(X^\omega_1,X^{\omega'}_1)|Z]]}
\end{align*}
where the joint distributions of $(Z,X^\omega_1,X^{\omega'}_1)$ and $(Z,X^\omega_2,X^{\omega'}_2, Y)$, respectively, are given by
\begin{align*}
    &Z \sim \mathbb{P}^{Z},\quad X_1^{\omega}|Z\sim \mathbb{P}^{X^{\omega}|Z},\quad X_1^{\omega'}|Z\sim \mathbb{P}^{X^{\omega}|Z},\quad X_1^\omega\indep X^{\omega'}_1|Z;\\
    &(Y,Z) \sim \mathbb{P}^{YZ},\quad X_2^{\omega}|Y,Z\sim \mathbb{P}^{X^{\omega}|Y,Z},\quad X_2^{\omega'}|Y,Z\sim \mathbb{P}^{X^{\omega}|YZ},\quad X_2^\omega\indep X^{\omega'}_2|X, Z.
\end{align*}

Then, the authors propose a graph-based estimator in \citet[sec.~3]{huang2022kernel}. Specifically, denote $\Ddot{Y}=(Y,Z)$ on $\mathcal{Y}\times \mathcal{Z}$, and let $G_n^Z$ and $G_n^{\Ddot{Y}}$ be graphs of a geometric graph functional, such as knn graphs, on $\mathcal{Z}$ and $\mathcal{Y}\times \mathcal{Z}$, respectively, with nodes $Z^n=(Z_1,\hdots,Z_n)$ and $\Ddot{Y}^n$, respectively. Moreover, let $\mathcal{E}(G_n^Z)$ and $\mathcal{E}(G_n^{\Ddot{Y}})$ denote the corresponding edge sets and $d_i^Z$ and $d_i^{\Ddot{Y}}$ be the degrees of $Z_i$ and $\Ddot{Y}_i$, $i=1,\hdots,n$. Then, the estimator of the KPC \eqref{equ:test_stat_KPC} called graph KPC is defined as in \citet[sec.~3.1]{huang2022kernel} by
\begin{align}\label{equ:graph_KPC}
    T_{\theta_{KPC}}^\omega((X^{\omega}, Y, Z)^n)=\frac{\frac{1}{n}\sum\limits_{i=1}^n \frac{1}{d_i^{\Ddot{Y}}}\sum\limits_{j:(i,j)\in\mathcal{E}(G_n^{\Ddot{Y}})}k_\mathfrak{X}(X_i^\omega,X_j^\omega)-\frac{1}{n}\sum\limits_{i=1}^n \frac{1}{d_i^{{Z}}}\sum\limits_{j:(i,j)\in\mathcal{E}(G_n^{{Z}})}k_\mathfrak{X}(X_i^\omega,X_j^\omega)}{\frac{1}{n}\sum\limits_{i=1}^n k_\mathfrak{X}(X_i^\omega,X_j^\omega)- \frac{1}{n}\sum\limits_{i=1}^n \frac{1}{d_i^Z} \sum\limits_{j:(i,j)\in\mathcal{E}(G_n^Z)} k_\mathfrak{X}(X_i^\omega,X_j^\omega)}
\end{align}
where $\theta_{KPC}=(k_\mathfrak{X}, k)$ with the kernel $k_\mathfrak{X}$ and its parameters as well as the number of nearest-neighbours $k$. 

While we could interchange $X^\omega$ and $Y$ in the KPC owing to the symmetry of CI \citep{dawid1979conditional}, and this can lead to different results since conditional dependence measures are not necessarily symmetric, our simulations indicate that the proposed order gives superior performance. This outcome may be attributed to the improved performance of the geometric graphs on the lower dimensional space $\mathcal{Y}\times \mathcal{Z}$ to the graphs for $(X^\omega, Z)$ on the higher dimensional $\mathfrak{X}\times \mathcal{Z}$. 

The CPT-KPC-based CIT (CPT-KPC) depends on $\theta_{CPT,KPC}=(\theta_{CPT},\theta_{KPC})$ where $\theta_{CPT}$ consists of the choice and the parameters of the learning model approximating $\mathbb{P}^{Y|Z}$ and the number $M$ of constructed approximate CI samples in the CPT, and $\theta_{KPC}$ consists of the choice of the kernel $k_\mathfrak{X}$ and its parameters, and the number of nearest neighbours in the knn graphs. 
The KPC accounts for the dimension of the feature representations by embedding them through a flexibly chosen kernel. In addition, it would even allow for non-Euclidean feature representations such as shapes by the use of a suitable kernel function on $\mathfrak{X}\times \mathfrak{X}$, as well as non scalar responses $Y$ and confounders $Z$ as long as they are elements of metric spaces. 
Nevertheless, the kernel $k_\mathfrak{X}$ affects the graph KPC, which is a common problem for kernel-based conditional dependence measures, see for example \citet{muandet2017kernel}, and thus, can decrease the power of the CPT-KPC. To study this dependence on the kernel, we analyzed the results for several kernels in a smaller simulation study based on the implementation of the KPC\footnote{\url{https://github.com/zh2395/KPC}}, and selected the Gaussian kernel since it performed best over multiple settings. 
In addition, the geometric graph functionals defined on $\mathcal{Z}$ and $\mathcal{Y}\times \mathcal{Z}$ measure the proximity within the confounder as well as the response. However, the geometric graph functionals depend on the number of nearest neighbors selected, where 1-nn leads to less biased estimates of the KPC while a larger number of neighbors is recommended to increase the power if the KPC is used as test statistic. For example, \citet{lin2023boosting} examined the effect of the number of nearest neighbors on the power of unconditional independence tests, and \citet{huang2022kernel} recommend in their implementation to select $k\approx 0.05n$ for samples smaller than 1000  and then increase $k$ sublinearly in $n$ for their variable selection algorithm. Consequently, in the smaller study, we examined \texttt{Deep-CPT-KPC}s for varying numbers of nearest neighbors and selected $k=10$ to trade off between T1E control and power. 
Then, we evaluate the \texttt{Deep-CPT-KPC}s together with Euclidean knn graphs, since this was the recommended choice in \citet[Remark~7]{huang2022kernel}. 
This leads to a time complexity of $\mathcal{O}(kn\log n)$ for estimating the graph KPC in \eqref{equ:graph_KPC} \citep{huang2022kernel}. Together with the time complexity of the CPT consisting of the generalized additive model, this leads to $\mathcal{O}((M+1)kn\log n+ns^2)$ for CPT-KPCs.
In summary, the hypothesis tested by the CPT-KPC is only constrained compared to \eqref{test_hypotheses_embedding} through weak, in the kernel literature standard, assumptions on the RKHS $\mathcal{H}_\mathfrak{X}$ and the kernel $k_\mathfrak{X}$ \citep[Remark~2]{huang2022kernel}, or $\mathcal{H}_\mathcal{Y}$ and the kernel $k_\mathcal{Y}$ if we interchange $X^\omega$ and $Y$. This allows for T1E control over a large $\mathcal{P}_0^\omega$ and correspondingly large $\mathcal{P}_0$. Moreover, the guaranteed T1E excess bound of the CPT together with the consistency of the KPC seems promising in terms of T1E control as well as power of the CPT-KPC. Nevertheless, since either the kernel embedding of $X^\omega$ or $Y$ is targeted, the test will detect small changes in  $X^\omega|Y,Z$ compared to $X^\omega|Z$, or $Y|X^\omega,Z$ compared to $Y|Z$, respectively, better.
We call DNCITs based on the CPT together with the KPC \texttt{Deep-CPT-KPC}.

\subsection{The Kernel Conditional Independence Test and the Randomized Correlation Test}\label{app:rcot}

We describe briefly the kernel-based RCoT from \citet{strobl2019approximate} and its approximation of the KCIT from \citet{zhang2012kernel}. 
The RCoT approximates the hypothesis and test statistic of the KCIT through
\begin{align*}
    H_{0,RCoT}^\omega:\|\mathcal{C}_{AB|C}\|_{F}^2=0\quad vs. \quad H_{1,RCoT}^\omega:\|\mathcal{C}_{AB|C}\|_{F}^2>0 
\end{align*}
estimated by
\begin{align*}
    T_{\theta_{RCoT}}^\omega((X^{\omega}, Y, Z)^n))=n\mathrm{tr}(\widehat{\mathcal{C}}_{AB| C}\widehat{\mathcal{C}}_{AB| C}^\top)
\end{align*}
where 
\begin{align*}
    \mathcal{C}_{AB|C}&=\mathbb{E}[(A_i-\mathbb{E}[A|C])(B_i-\mathbb{E}[B|C])^\top]\\
    \widehat{\mathcal{C}}_{AB| C}&=\frac{1}{n-1}\sum_{i=1}^n[(A_i-\widehat{\mathbb{E}}(A|C))(B_i-\widehat{\mathbb{E}}(B|C))^\top]\\
    A_i&=\phi^{X^\omega}(X_i^\omega)=\{\phi^{X^\omega}_1(X_i^\omega),\hdots,\phi^{X^\omega}_a(X_i^\omega)\},\quad \phi^{X^\omega}_j(X_i^\omega)\in \mathcal{G}_{X^\omega},\forall j,\\
    B_i&=\phi^{Y}(Y)=\{\phi^{Y}_1(Y_i),\hdots,\phi^{Y}_b(Y_i)\}, \quad \phi^{Y}_k(Y_i)\in \mathcal{G}_Y, \forall k,\\
    C_i&=\phi^{Z}(Z_i)=\{\phi^{Z}_1(Z_i),\hdots,\phi^{Z}_c(Z_i)\},\quad \phi^{Z}_l(Z_i)\in \mathcal{G}_Z,\forall l,
\end{align*}  
$\mathcal{G}_{X^\omega},\mathcal{G}_Y,\mathcal{G}_Z$ are spaces set to be the support of the process $\sqrt{2}\cos(W^\top \cdot + B), W\sim \mathbb{P}^W, B\sim Unif([0,2\pi])$ with $\mathbb{P}^W$ set to a Gaussian distribution with standard deviations $\sqrt{\sigma_A^2/2}$,  $\sqrt{\sigma_B/2}$, and $\sqrt{\sigma_C/2}$, and $\|\cdot \|_F$ denotes the Frobenius norm. That means, $a, b$ and $c$ functions are drawn from the spaces $\mathcal{G}_{X^\omega},\mathcal{G}_Y$ and $\mathcal{G}_Z$, respectively. Moreover, the conditional expectations $\mathbb{E}(A|C)$ and $\mathbb{E}(B|C)$ are approximated with linear ridge regression solutions. Thus, $\theta_{RCoT}=(a,b,c,\sigma_A, \sigma_B, \sigma_C, \lambda)$ with $a,b,c$ for the number of Fourier functions of $A,B,C$, the smoothing parameters $\sigma_A,\sigma_B,\sigma_C$ of the radial basis function kernels and the regularization parameter $\lambda$ of the linear ridge regressions. 

This approximates KCIT's hypotheses and the corresponding test statistic in three ways:  1.) Replacing $\Ddot{X}^\omega=(X^\omega,Z)$ by $X^\omega$. 
This corresponds to a test of weak CI of the data mapped into RKHSs instead of strong CI, as discussed in \citet{li2020nonparametric}.
Nevertheless, RCoT is usually similar in its performance compared to the corresponding Randomized
Conditional Independence Test (RCIT) using $\Ddot{X}^\omega$, which was also developed in \citet{strobl2019approximate}; 
2.) The kernel maps of $\Ddot{X}^\omega$ and $Y$ are replaced by low dimensional, random Fourier features. This is based on the approximation of continuous shift-invariant kernels through low-dimensional Fourier features, which was shown to perform well on large-scale data sets \citep{rahimi2007random}; 3.) The conditional expectations $\mathbb{E}[A|Z]$ and $\mathbb{E}[B|Z]$ are approximated by linear ridge regressions on $C$, which implies an approximation of the empirical partial cross-covariance matrix $\widehat{\mathcal{C}}_{AB| Z}$ through $\widehat{\mathcal{C}}_{AB| C}$. It was shown in \citet{sutherland2015error} that, under appropriate assumptions, the linear ridge regression solutions converge with an exponential rate in $c$ to the corresponding kernel ridge regression, justifying this approximation.

Asymptotic distributions and corresponding approximations are derived for the test statistics of KCIT and RCoT. 
The RCoT depends on the hyperparameters $(a,b,c,\sigma_A, \sigma_B, \sigma_C, \lambda)$. As described in \citet{zhang2012kernel,strobl2019approximate}, $\lambda\approx 0.1$ is a reasonable choice for low-dimensional $Z$ as in our setting. Moreover, $\sigma_A, \sigma_B, \sigma_C$ are chosen empirically on the first 500 observations or in sample, if the number of observations is smaller than 500. Compared to other nonparametric CITs discussed later, these are relatively few hyperparameters and the test performed well in our simulations under these settings and the default parameters $a=5, b=5, c=100$.
The RCoT is faster than the KCIT due to the three approximations discussed previously, as well as an approximation of the asymptotic distribution under the null of $\widehat{\mathcal{C}}_{AB| C}$ by a Lindsay-Pilla-Basak approximation, which makes a comparison with the null hypothesis computationally inexpensive. In particular, the RCoT has a time complexity of $\mathcal{O}(((\dim_{X^\omega}+\dim_{Z})a+(\dim_{Y}+\dim_{Z})b+\dim_Zc+(a+b)c^2)n)$, which reduces to $\mathcal{O}(n)$ for fixed $a,b,c, \dim_{X^{\omega}}, \dim_Y,\dim_Z$ \citep[Proposition~6]{strobl2019approximate}. 
Besides that, the RCoT covers a wide range of functional relationships between the random objects due to their Fourier transformations. Additionally, the projection of $X^\omega$ into the lower dimensional space $\mathcal{G}_{X^\omega}$ reduces the dimension of the feature representations $X^\omega$, which is particularly advantageous in our setting and allows for relatively high-dimensional $X^\omega$. Therefore, in contrast to the recommendation in \citet{strobl2019approximate}, we typically aim for a larger number of Fourier features $a$ to capture most information in the high dimension of $X^\omega$. 
Furthermore, the test is stably implemented in R\footnote{\url{https://github.com/ericstrobl/RCIT}}.
Moreover, we note that for small sample sizes, it may be preferable to use the KCIT due to a failure of the RCoT in the T1E control in these settings due to the approximations, see for example the simulation studies in \citet{sen2017model, zhang2023conditional}. Furthermore, for more than twenty confounders, other CITs may be more appropriate, as discussed in \citet{sen2017model}. 
In summary, the hypotheses considered in the RCoT are restricted compared to the hypotheses \eqref{test_hypotheses_embedding} only by the relatively weak assumption on the kernel $k_{\mathfrak{\Ddot{X}}}k_\mathcal{Y}$ and the RKHS $\mathcal{H}_\mathcal{Z}$, as well as the restriction to approximate Gaussian kernels in $\mathbb{P}^W$ and the corresponding, well-founded approximations of the KCIT. 
We call the corresponding DNCIT $\texttt{Deep-RCoT}$. 

\subsection{Prediction-Based Conditional Independence Tests}\label{app:pred_cits}

We evaluate the applicability of prediction-based CITs to vector-scalar-valued data. We can apply them by testing for an increase in the prediction accuracy of $Y$ using $X^\omega,Z$ compared to the prediction solely with $Z$, removing the predictive information within $X^\omega$. This can be done with various approaches as reviewed in \citet{covert2021explaining} under the perspective of model explanations, and translates the CIT typically to a weaker mean dependence test, accounting only for parts of the conditional distribution of $Y$ encoded through the loss function used to fit the prediction model.
In the following, we describe the approaches developed explicitly for CI testing, namely the CPI \citep{watson2021testing}, the FCIT \citep{chalupka2018fast} and the PCIT \citep{burkart2017predictive}. 

Therefore, let $f\in \mathcal{M}_{\mathfrak{X}\mathcal{Z}\to\mathcal{Y}}, \mathcal{M}_{\mathfrak{X}\mathcal{Z}\to\mathcal{Y}}:\mathfrak{X}\times \mathcal{Z}\to \mathcal{Y}$ be a prediction function for the scalar $Y$ using the feature representations $X^\omega$ and the confounders $Z$. Then, the CPI removes the predictive information of $X^\omega$ on $Y$ by constructing knockoffs $\widetilde{X^\omega}$, which were introduced by \citet{barber2015controlling, candes2018panning}. Particularly, for the risk $R_L(f,(X^\omega,Y,Z))=\mathbb{E}[L(f,(X^\omega, Y,Z))]$ and a non-negative, real-valued loss function $L:\mathcal{M}_{\mathfrak{X}\mathcal{Z}\to\mathcal{Y}}\times (\mathfrak{X}\times \mathcal{Y}\times\mathcal{Z})\to \mathbb{R}_{\geq 0}$, the CPI is a measure of conditional dependence as
\begin{align*}
    CPI((X^\omega, Y,Z))=\mathbb{E}[L(f,(\widetilde{X^\omega}, Y,Z))-L(f,(X^\omega, Y,Z))]
\end{align*}
Then, they propose to test \eqref{test_hypotheses_embedding} by testing 
\begin{align*}
    &H_{0, CPI}^\omega:CPI((X^\omega, Y,Z))\leq 0 \iff R_L(f,(\widetilde{X^\omega},Y,Z))\leq R_L(f,(X^\omega,Y,Z))  \\
    vs. \quad &H_{1,CPI}^\omega:CPI((X^\omega, Y,Z))>0 \iff R_L(f,(\widetilde{X^\omega},Y,Z))> R_L(f,(X^\omega,Y,Z)). 
\end{align*}
The CPI can be estimated by the empirical risks leading to the corresponding test statistic
\begin{align*}
    T_{\theta_{CPI}}^\omega((X^\omega, Y,Z)^n)=\frac{1}{n_{test}}\sum_{i=1}^{n_{test}}L(f, (\widetilde{X^\omega}_i,Y_i,Z_i))-\frac{1}{n_{test}}\sum_{i=1}^{n_{test}}L(f, (X^\omega_i,Y_i,Z_i))
\end{align*}
over a test data set $(X^\omega, Y,Z)^{n_{test}}$ of size $n_{test}$, here obtained after splitting $(X^\omega, Y,Z)^n$ into train and test data set. Parameters $\theta_{CPI}$ consist of the knockoff construction parameters, the loss $L$ and the prediction model $f$ as well as its parameters.  

In contrast to the CPI and its use of knockoffs, the PCIT and FCIT remove $X^\omega$ from the prediction to obtain a prediction function $g\in \mathcal{M}_{\mathcal{Z}\to \mathcal{Y}}, \mathcal{M}_{\mathcal{Z}\to \mathcal{Y}}:\mathcal{Z}\to \mathcal{Y}$, by regressing $Y$ only on the confounder $Z$. Then, the authors propose to test \eqref{test_hypotheses_embedding} by testing 
\begin{align*}
    &H_{0, FCIT/PCIT}^{\omega}: R_L(g, (Y, Z))\leq R_L(f, (X^\omega, Y,Z)) \\
    vs. \quad &H_{1,FCIT/PCIT}^\omega:R_L(g, (Y, Z))> R_L(f, (X^\omega, Y,Z)),
\end{align*}
which can again be estimated by the corresponding empirical risks. While the FCIT implements a decision tree algorithm to learn $g$ and $f$, the PCIT allows to fit a set of regression and classification models as well as their ensembles. However, PCIT's implementation is currently unstable and not further used in our study.

For the comparison to the null hypothesis, any paired mean difference test such as the paired t-test or Fisher exact test can be used to obtain p-values. All three tests can rely on asymptotic normality results of the empirical risk estimates.  

We call the corresponding DNCITs \texttt{Deep-CPI} and \texttt{Deep-FCIT}. They reduce conditional independence to conditional variable importance. Thereby CI testing is translated to mean dependence or at least loss dependent CI testing, which has the advantage that it draws on a large literature of variable importance testing, while often being sufficient to detect deviations from the null. However, this means that the tests can only have power for data generating processes for which the difference between $\mathbb{P}^{Y|X^\omega,Z}$ and $\mathbb{P}^{Y|\widetilde{X^\omega},Z}$ or $\mathbb{P}^{Y|Z}$, respectively, is detectable by the chosen loss. Particularly, the tests focus on specific properties of the distribution encoded in the loss $L$ instead of the whole conditional distribution of $Y$, resulting in a loss of power for differences in other properties of the distribution. 
Additionally, it usually restricts the problem to a specific learner or an ensemble of learners and the conditional variable importance for that learner(s). Therefore, while the tests itself do not impose restrictions w.r.t. $X^\omega, Y, Z$ and their functional relationships, the tests impose these implicitly through the used learners. In particular, prediction-based DNCITs only account for functional relationships between $X^\omega$ and $Y$ which can be approximated by the learner. This leads to a potential loss in power if $f$ does not cover the true functional relationship between $X^\omega$ and $Y$, and an increase in the T1E if $g$ does not estimate the true functional relationship between $Y$ and $Z$ well enough or if the knockoffs are not well constructed, similar to the T1E excess for CPT-based CITs. 
On the one hand this flexibility in the choice of the learner allows for models adapted to a specific data setting. 
On the other hand, it is difficult to obtain a default CIT, which would ensure consistency over test results, making these tests rather hyperparameter dependent. 
Moreover, the sample split necessary to fit the learner results in a reduced power. Finally, an advantage can be that their implementations\footnote{FCIT in python: \url{https://github.com/kjchalup/fcit}; PCIT in python: \url{https://github.com/alan-turing-institute/pcit}; CPI in R: \url{https://github.com/bips-hb/cpi}} are based on often stable implementations of learners for the functions $f$ and $g$.
Particularly, the CPI-based CIT has a time complexity of $\mathcal{O}(ue+v+w)$ where $u$ is the complexity of the learner $f$, $e$ of the empirical risk estimator, $v$ of the knockoff sampler and $w$ of the inference procedure such as the t-test \citep{watson2021testing}. 
The FCIT has a time complexity, without the inclusion of the knockoff but with additional costs to fit the learner $g$,  of $\mathcal{O}(n^2\log n (p+q))$ \citep{chalupka2018fast}.

As a final remark on prediction-based CITs, we note that they could be applied directly to the images, either by removing $X$ in a model refit as in FCIT and PCIT, or by using knockoffs for $X$.
However, this exacerbates the challenges described in the previous paragraph. That is, T1E, power, and complexity are affected as follows: First, the T1E increases for missconstruction of the knockoffs of $X$, which are harder to construct than knockoffs of $X^\omega$. 
Second, the power decreases when the model cannot account for in increase in predictive accuracy using $X$ and $Z$ compared to the predictive accuracy using $Z$ alone. While there exist many models to estimate the additional predictive accuracy using $X^\omega$ and $Z$, often working well with established automated hyperparameter tuning, the models such as neural networks for $X$ and $Z$ are often strongly dependent on their hyperparameters without simple tuning methods, making the performance of the tests even stronger hyperparameter dependent and difficult to fit on small data sets. Finally, the models for learning $f$ increase in complexity, which greatly increases the time complexity for each test. This makes the corresponding tests computationally expensive or even prohibitive, especially for multiple tests including the images, compared to DNCITs using the same feature representations over all multiple tests. Therefore, we do not investigate this approach further in this paper.

\subsection{The Conditional Mutual Information K-Nearest-Neighbour Test}

The metric-based CMIknn CIT of \citet{runge2018conditional} is based on the CMI as a test statistic, testing 
\begin{align*}
    &H_0^{CMIknn}:CMI((X^\omega,Y,Z))=\int\int\int dxdydz p(x,y,z)\log\frac{p(x,y|z)}{p(x|z)p(y|z)}=0 \\
    vs. \quad &H_1^{CMIknn}:CMI((X^\omega,Y,Z))>0,
\end{align*}
since $CMI((X^\omega,Y,Z))=0$ if and only if $X^\omega\indep Y|Z$ \citep{runge2018conditional}. Then, the authors present the estimator 
\begin{align*}
    T^{CMIknn}((X^\omega,Y,Z)^n)=\psi(k_{CMI})+\frac{1}{n}\sum_{i=1}^n\left[\psi(k_i^z)-\psi(k_i^{xz})-\psi(k_i^{yz})\right],
\end{align*}
where $\psi(x)=\diff{}{x}\ln \Gamma(x)$, $\Gamma(\cdot)$ the Gamma function, $k_{CMI}$ is a pre-specified number of nearest neighbors in $\mathfrak{X}\times \mathcal{Y}\times \mathcal{Z}$
defining implicitly the local distances $\epsilon_i, i=1,\hdots, n$
and $k_i^z,k_i^{xz}, k_i^{yz}$ are the corresponding numbers of nearest neighbors for $i=1,\hdots,n$ in $\mathcal{Z}, \mathfrak{X}\times \mathcal{Z}$ and $\mathcal{Y}\times \mathcal{Z}$, respectively, with distance strictly smaller than $\epsilon_i$. 

This test statistic is compared to the null hypothesis through a local permutation scheme. In particular, the $k_{perm}$-nearest neighbors in $\mathcal{Z}$ are computed and for each $Y_i,i=1,\hdots,n$ its neighbours are selected by these nearest neighbours in $\mathcal{Z}$. Then, each $Y_i$ in $Y^n$ is locally, randomly permuted $M$ times such that each $Y_i$ is, if possible, only used once in the permuted $(Y^{(m)})^n, m=1,\hdots,M$. Finally, the p-value is obtained through $p^{\omega}=M^{-1}\sum_{m=1}^M\mathds{1}\{\widehat{T}^{CMIknn}((X^\omega,Y^{(m)},Z)^n)\geq \widehat{T}^{CMIknn}((X^\omega,Y,Z)^n)\}$. 

The CMIknn CIT\footnote{CMIknn: \url{https://github.com/jakobrunge/tigramite}} is applicable to arbitrary dimensions of $X^{\omega},Y,Z$, which makes it applicable to the feature representations of the images and even more generally, metric spaces $\mathfrak{X},\mathcal{Y},\mathcal{Z}$. It depends only on the hyperparameters $k_{CMI}$ and $k_{perm}$, where the authors suggest a low, fixed $k_{perm}\in\{5,\hdots,10\}$ and a rule-of-thumb of $k_{CMI}\approx 0.1n \text{ or } 0.2n$. The T1E and power increase for larger $k_{perm}$, while it is theoretically guaranteed only for $k_{perm}=1$ that $(X^\omega,Y^{(m)},Z)^n, m=1,\hdots, M$ are approximately sampled from $\mathcal{P}_0^{\mathfrak{X}}$ \citep{sen2017model}. The power of the test increases in $k_{CMI}$ up to some maximum in the existing simulation studies in \citet{runge2018conditional}. The test has a computational cost of $\mathcal{O}(n^2)$ for the search for nearest neighbors in $\hat{T}^{CMIknn}$ and the permutation scheme, and scales roughly linearly in $k_{CMI}$ and $(p+q+1)$, where $k_{CMI}$ is the number of nearest neighbors used to estimate $T^{CMIknn}$ and $p+q+1$ is the dimension of $(X^\omega,Y,Z)$ \citep{runge2018conditional}. We call the corresponding DNCIT \texttt{Deep-CMIknn}.

\end{document}